**Bioinspired underwater soft robots: from biology to robotics and back**

*Lei Li, Boyang Qin, Wenzhuo Gao, Yanyu Li, Yiyuan Zhang, Bo Wang, Shihan Kong, Jian Wang, Dekui He, Junzhi Yu**

L. Li
Institute of Ocean Research, Peking University, Beijing, China.

L. Li, B. Y. Qin, W. Z. Gao, B. Wang, S. H. Kong, J. Z. Yu
College of Engineering, Peking University, Beijing, China.
E-mail: yujunzhi@pku.edu.cn

Y. Y. Li, D. K. He
Institute of Hydrobiology, Chinese Academy of Sciences, Wuhan, China

Y. Y. Zhang
College of Design and Engineering, National University of Singapore, Singapore, Singapore

J. Wang
Institute of Automation, Chinese Academy of Sciences, Beijing, China

**Abstract**

The ocean’s vast unexplored regions and the diversity of soft-bodied marine organisms have sparked a surge of interest in bioinspired underwater soft robotics. Drawing from biological adaptations such as remora adhesion, shark-skin drag reduction, and octopus-arm dexterity, recent robotic advances are expanding the capabilities of aquatic exploration. However, most studies remain unidirectional as they translate biological principles into robotic designs while often overlooking how robotics can, in turn, inform biological understanding. Here, we propose a holistic, bidirectional framework that integrates biological feature extraction (morphology, kinematics, sensing), robotic implementation (materials, actuation, control), and biological validation (hydrodynamics, adhesion mechanisms). We demonstrate that soft robots can serve as experimental tools to probe biological functions and test evolutionary hypotheses, including those in paleobiology. Their inherent compliance enables them to outperform rigid systems in unstructured environments, with applications in marine exploration, manipulation, and medicine. Looking forward, we introduce “biouniversal-inspired robotics”—a paradigm that transcends species-specific mimicry by identifying convergent principles across taxa to enable scalable, adaptable robotic design. Despite rapid progress, persistent challenges remain in material robustness, actuation efficiency, autonomy, and intelligence. By bridging biology and engineering, soft robots can both advance ocean technology and deepen scientific discovery.

## 1. Introduction

Covering over 70% of Earth's surface, the ocean hosts an extraordinary abundance of soft-bodied organisms that thrive across diverse habitats—from tranquil coral reefs and turbulent coastal zones to deep-sea trenches where pressures exceed 100 MPa. These organisms exhibit remarkable mechanical adaptability, enabling survival in environments characterized by strong currents, spatial constraints, and extreme pressure. To operate effectively in such conditions, underwater robots must overcome similar challenges. Traditional underwater platforms—autonomous underwater vehicles (AUVs) and remotely operated vehicles (ROVs)—rely on rigid materials and motor-driven systems to perform tasks such as seafloor mapping, deep-sea organism sampling, and environmental monitoring[1,2]. However, their limited maneuverability, low hydrodynamic efficiency, and potential to disrupt delicate ecosystems constrain their use in complex settings. These limitations have fueled the rise of soft robotics, which leverages flexible materials, distributed actuation, and bioinspired architectures to enable safer, more adaptable, and energy-efficient interaction with underwater environments[3].

Nature's evolutionary processes provide a powerful blueprint for robotic design. Over millions of years, marine organisms have evolved remarkable adaptations: remora fish use vertically aligned collagen fibers in their adhesive discs to enhance attachment strength[4]; shark skin's denticles reduce drag by controlling turbulent boundary layers[5]; octopuses manipulate objects with exceptional dexterity using soft, flexible arms[6]; and fish achieve high propulsion efficiency through coordinated undulation of their bodies and caudal fins[7]. These biological systems, particularly those involving soft materials, adaptable structures, and efficient locomotion, provide valuable inspiration for bioinspired underwater soft robots.

By mimicking natural materials (composite fibrous tissues[8], mussel byssus protein[9]), biological structures (octopus arms[10], seal whiskers[11]), locomotion strategies (fish swimming[12], jet propulsion[13], leg-kicking[14], and benthic locomotion[15]), sensing mechanisms (aquatic-type vision[16], lateral line system[17]) and collective behaviors (fish school[18], group foraging in manta rays[19]), bioinspired soft robots can achieve enhanced functionality in aquatic environments. This biologically grounded approach not only overcomes the limitations of rigid robots but also unlocks new possibilities for operations in sensitive and complex underwater settings. The evolution of bioinspired robotics typically progresses from superficial mimicry to deep functional emulation of biological systems, enabling increasingly sophisticated designs.

Recent advances in bioinspired soft robotics have been widely reviewed, covering materials, design, fabrication, actuation, modeling, sensing, control, and applications[3,20–54]. Notably, Ye et al.[21], Sarker et al.[22], and Coyle et al.[43] have provided comprehensive overviews; Yang et al.[25] have explored biologically derived design principles; Van Laake et al.[23] have focused on bioinspired autonomy; Ahmed et al.[31] have reviewed actuation methodologies. While both terrestrial and underwater bioinspired robots share core principles, underwater robots must contend with distinct challenges in hydrodynamics, waterproofing, communication, actuation, sensing, and locomotion. Despite the field's rapid growth, reviews specifically addressing underwater bioinspired soft robots remain limited[55–58]. For example, Qu et al.[55] reviewed advances in materials, actuation, and microswimming robots; Li et al.[56] focused on robots for deep-sea exploration; Youssef et al.[57] examined bioinspired designs and control strategies; and Hermes et al.[58] surveyed actuation strategies for crawling and swimming.

However, few reviews have systematically addressed the iterative process of translating biological principles into robotic systems and how these robots, in turn, contribute to biological research. This gap highlights the need for a holistic framework that fosters a bidirectional exchange of knowledge between biology and robotics, ultimately advancing both disciplines.

This review presents a comprehensive research paradigm for bioinspired underwater soft robotics, highlighting a bidirectional flow of knowledge—from biology to robotics, and from robotics back to biology. This perspective not only addresses fundamental scientific questions but also supports the development of robotic systems with adaptive capabilities resembling those of living organisms (**Figure 1**). The structure of the review is as follows: We begin by extracting key biological features across materials, morphology, kinematics, sensing, and swarm strategies. These insights guide the design, fabrication, and implementation of soft robotic systems. We then explore how these robots serve as platforms to investigate biological and paleobiological mechanisms. Building on this foundation, we review current applications and outline future directions, including the concept of biouniversal-inspired robotics. By adopting an interdisciplinary approach, this framework has the potential to deepen our understanding of biological systems while advancing the development of underwater robots, benefiting both scientific discovery and practical applications.

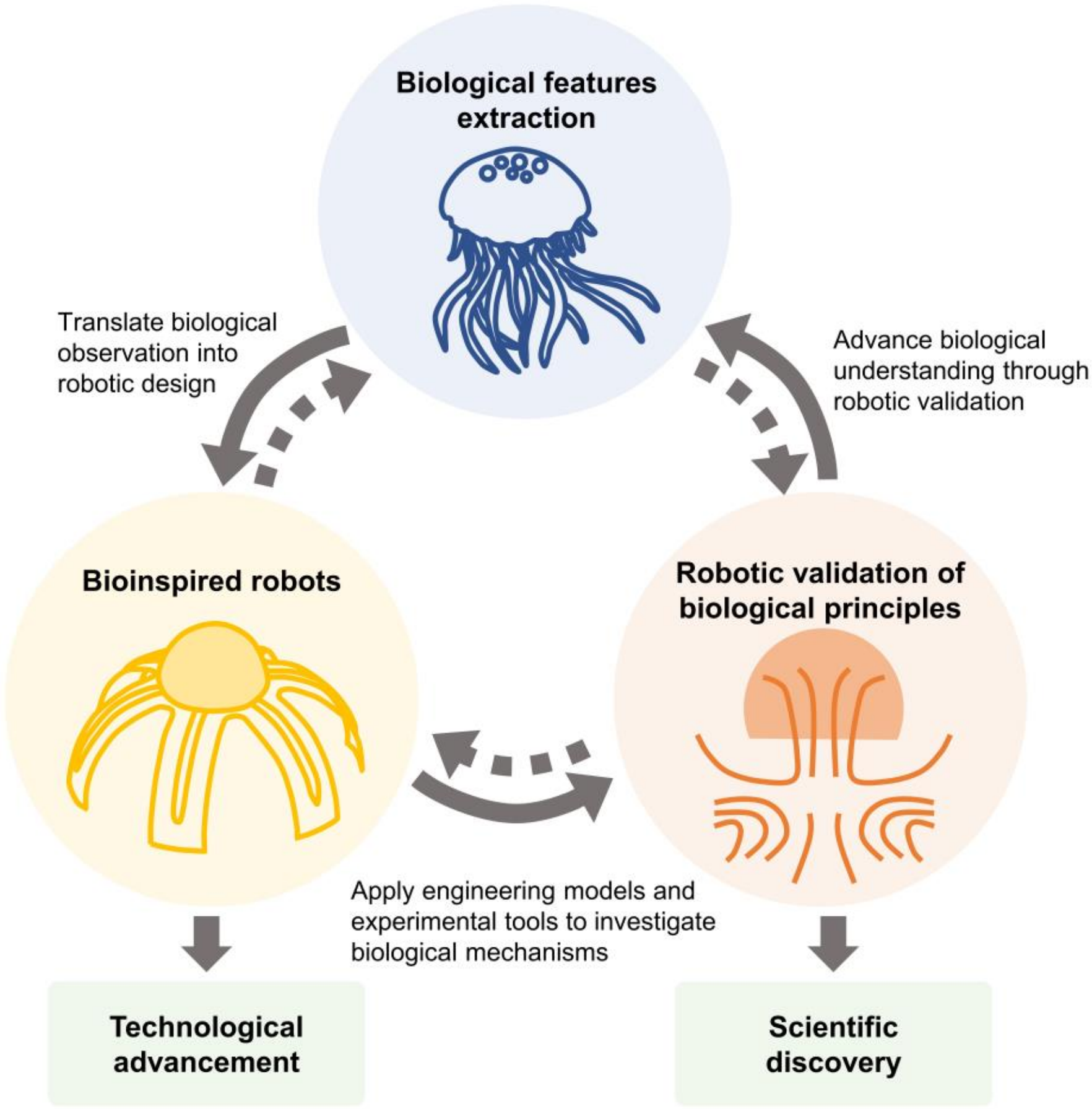

**Figure 1. Framework for bioinspired underwater soft robotics, illustrating the bidirectional integration of biological principles and robotic implementation.**

## 2. Biological features extraction

Aquatic organisms offer a rich source of inspiration for underwater soft robotics. Echinoderms such as starfish regenerate lost tissues via pluripotent stem cell clusters at the base of each arm[59]. Within vertebrates, tuna utilize muscle-tendon energy transmission for high-speed propulsion[60], while paddlefish achieve 95% plankton filtration efficiency using branchial rakers spaced at 50 μm[61]. Boxfish sense complex 3D flow fields through lateral line neuromasts[62], and remoras employ suction discs on their heads to achieve stable attachment to host surfaces[63]. In extreme environments, hadal snailfish withstand hydrostatic pressures exceeding 110 MPa through decentralized skeletal systems and pressure-diffusing soft tissues[64]. Cnidarian jellyfish achieve low-energy propulsion through bell-driven vortex streets[65], while cephalopods such as octopuses manipulate hyper-redundant arms equipped with over 200 suckers each[66]. Marine mammals like dolphins discriminate fine-scale targets using biosonar operating between 20-150 kHz, with millimeter-level resolution[67]. These adaptations, refined over 450 million years of evolution—spanning from mussel foot protein-based self-healing adhesion[68] to sardine shoals' rapid collective maneuvers[69]—form a diverse biological database. By systematically extracting pivotal principles across material properties, morphology, locomotion, environmental adaptation, sensing, and swarm intelligence, these insights propel advancements in biomimetic, bioinspired, and bionic underwater soft robotics.

### 2.1 Material features

Aquatic organisms have evolved soft materials with adaptive mechanical properties, self-healing abilities, and exceptional environmental resilience through structural optimization. For example, marine mussels secrete byssus threads that provide strong underwater adhesion and self-healing, facilitated by multiscale interfacial anchoring (hydrogen bonding/metal chelation) and dynamic $Fe^{3+}$ crosslinking[70]. The remora suction disc lip, composed of stratified epithelium (epithelium layer), cross-fiber collagen (dermal layer), and vertically aligned collagen fibers (core layer), functions as a composite structure material, balancing flexibility and strength[4,71].

Characterizing these biomaterials requires an integrative approach combining mechanics, chemistry, and functional analyses[72,73]. Mechanical testing—including tensile, compression, and nanoindentation—quantifies elastic modulus (0.1–100 MPa) and viscoelasticity. Dynamic thermomechanical analysis (DMA) quantifies temperature- and frequency-dependent changes in storage modulus. Chemical and structural analyses provide complementary insights. Fourier-transform infrared spectroscopy (FTIR) identifies molecular bonds and functional groups; X-ray diffraction (XRD) determines crystalline phases and lattice structure; X-ray photoelectron spectroscopy (XPS) reveals surface elemental composition and bonding states, including C/O/N ratios. Functional assays evaluate biodegradability, swelling kinetics, and biocompatibility.

A notable example is the study by Abdon et al.[74] on squid-derived self-healing materials from *Loligo vulgaris* suction cups (**Figure 2A**). Using FTIR, XRD, and cross-species sequence analysis, they identified unique tandem repeat (TR) motifs in the structural proteins. These TR sequences, combining disordered flexible segments and β-sheet crystalline regions, enable high strength and rapid self-repair through thermally activated hydrogen bond reorganization.

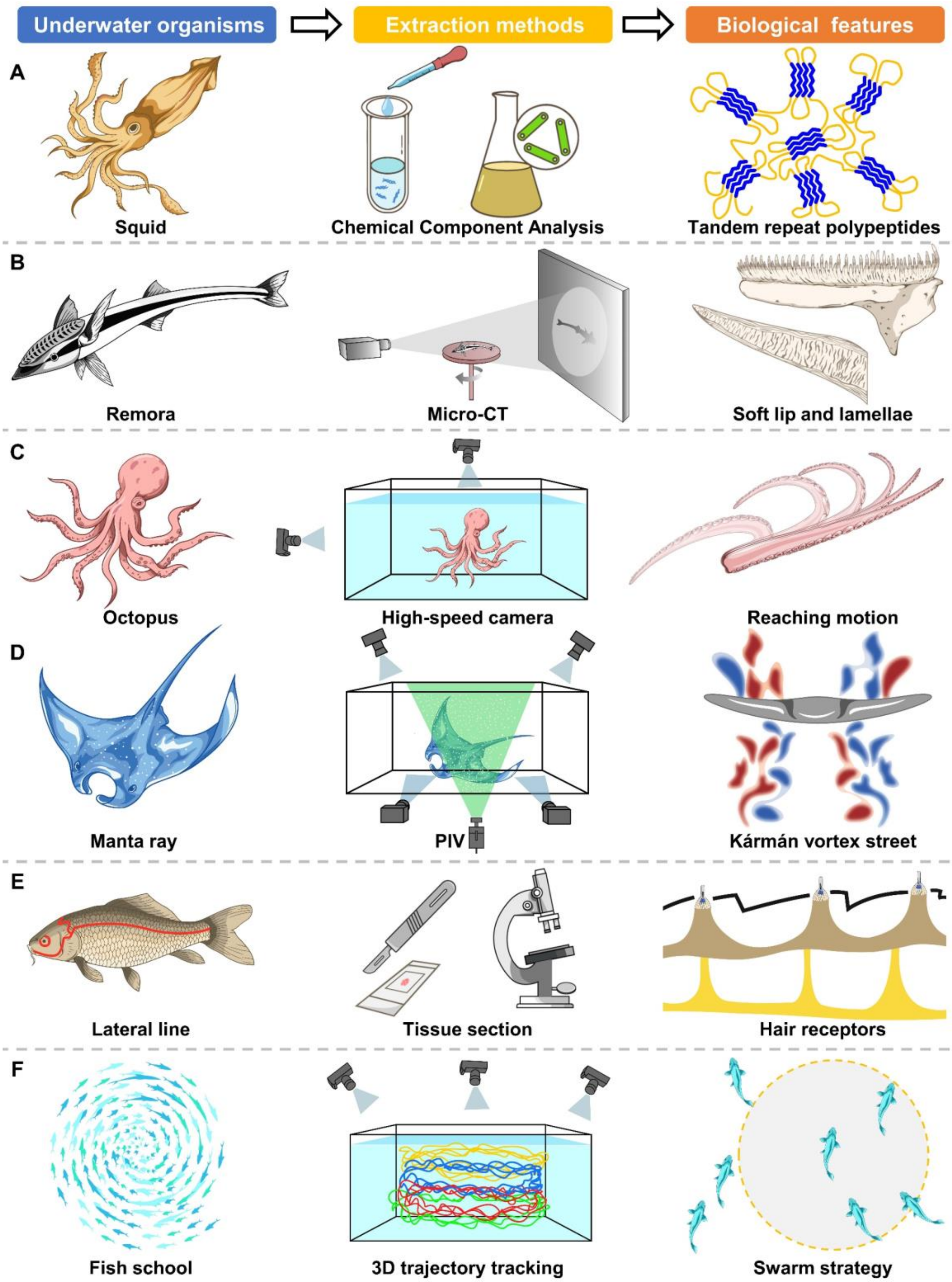

**Figure 2. Extraction of typical biological features, illustrating the process from biological**

**systems to identified characteristics through extraction methods.** (A) Extraction of material characteristics from the self-healing protein in squid suction cups, modified from Abdon et al., 2020[74]. (B) Extraction of morphological features from the remora adhesion disc, including the soft lip, lamellae and spinules, modified from Wang et al., 2017[75] and Li et al., 2022[76]. Extraction of kinematic properties from octopus arm reaching locomotion (C) and tuna tail-swing hydrodynamics (D), modified from Xie et al., 2023[77] and Gao et al., 2023[78], respectively. (E) Extraction of sensing capabilities from the fish lateral line, modified from Peleshanko et al., 2007[79]. (F) Extraction of swarm behaviors from fish schooling behavior, modified from Berlinger et al., 2021[80].

### 2.2 Morphological features

Aquatic organisms have evolved diverse morphological adaptations in response to environmental pressures such as hydrostatic forces, flow gradients, and light conditions. These include optimized body shapes, specialized surface textures, and distinct appendages. For instance, shark skin features densely microstructured denticles with 3-5 asymmetrical longitudinal ridges, reducing drag and enhancing swimming efficiency[81–83]. Octopus arms exhibit a complex muscular architecture, where transverse muscles (perpendicular to the axis) surround the central nerve cord, while longitudinal and helical oblique muscles, supported by elastic collagen fibers, enable elongation, shortening, and torsional control under volume conservation[66]. Sea turtles, such as green turtles and hawksbills, possess streamlined flippers for propulsion and robust skeletons for terrestrial movement[84]. Fish fins, composed of flexible membranes supported by fin rays, play crucial roles in locomotion and balance. The caudal fin, with shapes like rounded, forked, or lunate, drives propulsion, while dorsal, anal, and pelvic fins stabilize and steer[85,86].

Morphological analysis requires multi-scale imaging techniques[87,88]. Histochemical staining (hematoxylin-eosin, Masson's trichrome) and fluorescent labeling visualize tissue structures and developmental dynamics. Confocal laser scanning microscopy (CLSM) provides high-resolution 3D imaging of labeled structures like muscle fibers and neural networks in live or fixed specimens. Electron microscopy techniques, including scanning (SEM), cryo-electron (cryo-EM), and transmission (TEM), reveal nanoscale surface textures, ultrastructures, and intracellular details. Micro-CT and synchrotron X-ray imaging allow non-destructive 3D reconstruction of skeletal systems at submicron resolution. Real-time imaging tools like high-frequency ultrasonography and confocal microscopy capture dynamic soft tissue behaviors, such as swim bladder volume changes, while AI-driven image analysis automates the high-throughput extraction of evolutionary traits.

The remora (*Echeneidae*) adhesion disc exemplifies integrated macro- and micro-morphology (Figure 2B). Wang et al.[75] and Li et al.[76] used histochemical staining, micro-CT, and SEM to reveal complex components including soft tissue and bony structure of lamellae, spinules, connective tissue, a disc lip, muscles for actuating the lamellae, and blood vessels. The soft connective tissue connects the lamellae and the soft lip, and the flexible lamellae have a tilted angle on top. These complex morphological features underpin remora's stable attachment and serve as a foundation for biomimetic research.

**2.3 Kinematic features**

Kinematic adaptations in aquatic species reflect the optimization of energy transfer, movement efficiency, and environmental responsiveness. By integrating fluid dynamics, models link morphological parameters (curvature, stiffness, surface features) with flow characteristics (vortex evolution, pressure gradients, boundary layers). This reveals the synergistic mechanisms that maximize propulsion efficiency and minimize energy dissipation through flexible deformation-fluid interactions. For instance, jellyfish propulsion relies on passive energy recovery: during contraction, a starting vortex forms, while relaxation generates a stopping vortex, creating a high-pressure zone that enhances thrust and increases pulsation distance by 30%[89]. The dorsal fin oscillations of the bluegill sunfish (*Lepomis macrochirus*) generate downstream thrust and lateral forces, which interact with the tail fin's flow field to improve propulsion efficiency by 15%[90]. Tuna utilizes periodic tail oscillations to generate a reverse Kármán vortex street, shedding alternating vortices that oppose the incoming flow and enhance propulsion efficiency[91].

Studying aquatic motion requires advanced kinematic and flow visualization tools[92,93]. High-speed video analysis captures movement trajectories, multi-angle imaging reconstructs 3D motion, and motion capture systems track precise limb kinematics. Particle image velocimetry (PIV) visualizes velocity fields by adding micro-particles to water and analyzing flow interactions. Wireless sensors provide real-time movement data, computational fluid dynamics (CFD) simulates locomotion patterns, and bioelectrical measurements reveal neural control mechanisms. These methods offer a comprehensive understanding of aquatic motion.

Two distinct kinematic models illustrate these principles. Octopuses, with its soft, boneless arms, employs the bend propagation strategy to reach a target object (Figure 2C)[77]. Bend formation and propagation result from precise, coordinated contractions of transverse and longitudinal muscles. As the bend travels distally, the proximal arm stiffens to provide support[66]. In contrast, manta rays achieve propulsion by synchronizing pectoral fin undulations, generating 3D vortex rings and a reverse Kármán vortex street that shed coherent vortices to reduce flow resistance, improve lift-to-drag ratio, and enhance efficiency (Figure 2D)[78].

**2.4 Sensing features**

Aquatic animals rely on highly specialized sensing systems to perceive complex underwater environments. These include optical (eyes), mechanical (lateral line, whiskers), chemical (olfactory, taste), electromagnetic (skin), and acoustic (ears) sensors[94]. For instance, the lateral line hair cells of larval zebrafish (*Danio rerio*) detect shear forces via cilia bundles (10-15 μm), enabling precise vorticity sensing[95]. Additionally, zebrafish have a unique olfactory system called the main olfactory system, consisting of a single olfactory epithelium that functions as the primary site for odor detection[96]. Sharks and rays use the ampullae of Lorenzini, filled with highly conductive gel, to detect weak electric fields for prey localization[97]. The Mexican cavefish (*Astyanax mexicanus*) produces “sharp click” sounds, used for aggression in surface-dwelling populations and foraging cues in cave-dwelling variants[98].

Multiscale sensory analysis requires integrated techniques, including Sudan red staining and tissue clearing for structural visualization, light-sheet microscopy for large-scale neurovascular mapping, micro-CT for nondestructive imaging, chemical digestion for soft tissue exposure (pancreatic enzyme digestion to expose neural ganglia), and focused ion beam scanning electron microscopy (FIB-SEM) for nanoscale synaptic analysis. Calcium imaging further reveals real-time neural activity, enabling a comprehensive understanding of sensory transduction mechanisms [99].

An exemplary case is the blind cavefish, which has adapted to darkness by enhancing lateral line sensitivity (Figure 2E). Increased lateral line ganglia density and a glycoprotein gel cap (cupula) amplify weak hydrodynamic signals, allowing detection of low-frequency vibrations (3-5 kHz) and water flows as slow as 0.075 mm/s. These adaptations ensure precise environmental perception even in complete darkness[79].

**2.5 Swarm strategies**

Underwater swarms, such as fish schools, exhibit self-organizing behaviors based on three critical rules: short-range repulsion (collision avoidance), mid-range alignment (movement synchronization), and long-range attraction (group cohesion). These rules enable formations such as polarized swarms or vortex rings[100]. The swarm adapts to threats (dispersing to escape predators) and optimizes foraging (grouping to gather plankton). Local information is shared through visual and lateral line perception, enabling rapid, real-time coordination without a central controller. Manta rays, for instance, adopt distinct foraging strategies: chain foraging (alignment with food flow), cyclone foraging (vortex formation to concentrate plankton), and somersault foraging (rolling to enhance food capture)[101].

Studying underwater swarm behavior relies on multimodal sensing and cross-scale modeling[102–104]. Acoustic tracking (multibeam sonar) maps 3D swarm trajectories in real time, distinguishing movement patterns via time-frequency analysis. Optical imaging (high-speed cameras on autonomous underwater vehicles) captures fine-scale interactions like cooperative hunting, though visibility is limited by turbidity. Biotelemetry (tags with accelerometers and heart rate sensors) links individual physiology to environmental conditions, revealing energy trade-offs in swarm dynamics. Computational models, such as individual-based models (IBMs) and fluid-dynamics-coupled simulations, predict optimal migration strategies, while deep learning (3D convolutional neural networks) identifies novel decision-making patterns from large datasets.

Many fish species, such as damselfish and sergeant major fish, rely on their visual system to perceive the position and movement of nearby individuals. They achieve real-time behavioral coordination by recognizing species-specific patterns, such as stripes in zebrafish or bioluminescence in deep-sea fish. Schooling fish self-organize without central control, using local interactions to form synchronized swimming patterns (Figure 2F)[80]. In dynamic formations, such as circular swimming, fish adjust spacing and synchronize movement phases (tailbeat frequency) to reduce group swimming resistance, thus significantly lowering overall energy consumption[105].

The biological features of aquatic organisms provide a comprehensive foundation for the design of biomimetic underwater soft robots. By systematically extracting essential features from materials, morphology, kinematics, sensing, and swarm behavior, researchers can develop adaptive, efficient, and multifunctional robotic systems.

## 3. Design and implementation of bioinspired underwater soft robots

Building on the biological features identified in Section 2, the design of bioinspired underwater soft robots involves the integration of diverse traits to achieve functional adaptability. Key considerations include the mechanical behavior of flexible materials (elasticity and viscoelasticity of silicone), the geometric topology of biomimetic structures (muscular fiber arrangements in octopus' arms), and the coordination of actuation with movement (timing control of pneumatic artificial muscle). This section explores how interdisciplinary strategies translate biological insights into programmable, controllable soft robots and outlines essential manufacturing technologies, including material selection, actuation integration, structural formation, sensing, control, and power.

### 3.1 Soft materials

Material stiffness is commonly evaluated by Young's modulus. Conventional engineering materials, such as metals and rigid plastics, exhibit moduli above $10^9$ Pa, whereas rubbers and biological tissues, including skin and muscle, typically range between $10^4$ and $10^9$ Pa. While the distinction between "soft" and "hard" is relative, materials with a modulus below $10^9$ Pa are generally classified as soft.

The bodies of biomimetic underwater soft robots are primarily constructed from compliant materials, whose flexibility and large deformation capabilities allow them to replicate the motion patterns of aquatic organisms. Key material categories include elastomers, hydrogels, and smart responsive materials (**Table 1**). Elastomers, such as polydimethylsiloxane (PDMS), silicone rubber, and liquid crystal elastomers, are widely used due to their high elasticity, underwater chemical stability, and ease of processing. Hydrogels, with high water content and excellent biocompatibility, are well-suited for mimicking soft tissues. Smart responsive materials, such as shape memory polymers, can respond to environmental stimuli, while liquid metals and electroactive polymers offer exceptional conductivity and electromechanical properties, enabling advanced underwater sensing and actuation. These material systems form the foundation for constructing underwater robots that are not only mechanically compliant but also capable of responding dynamically to their operating environment.

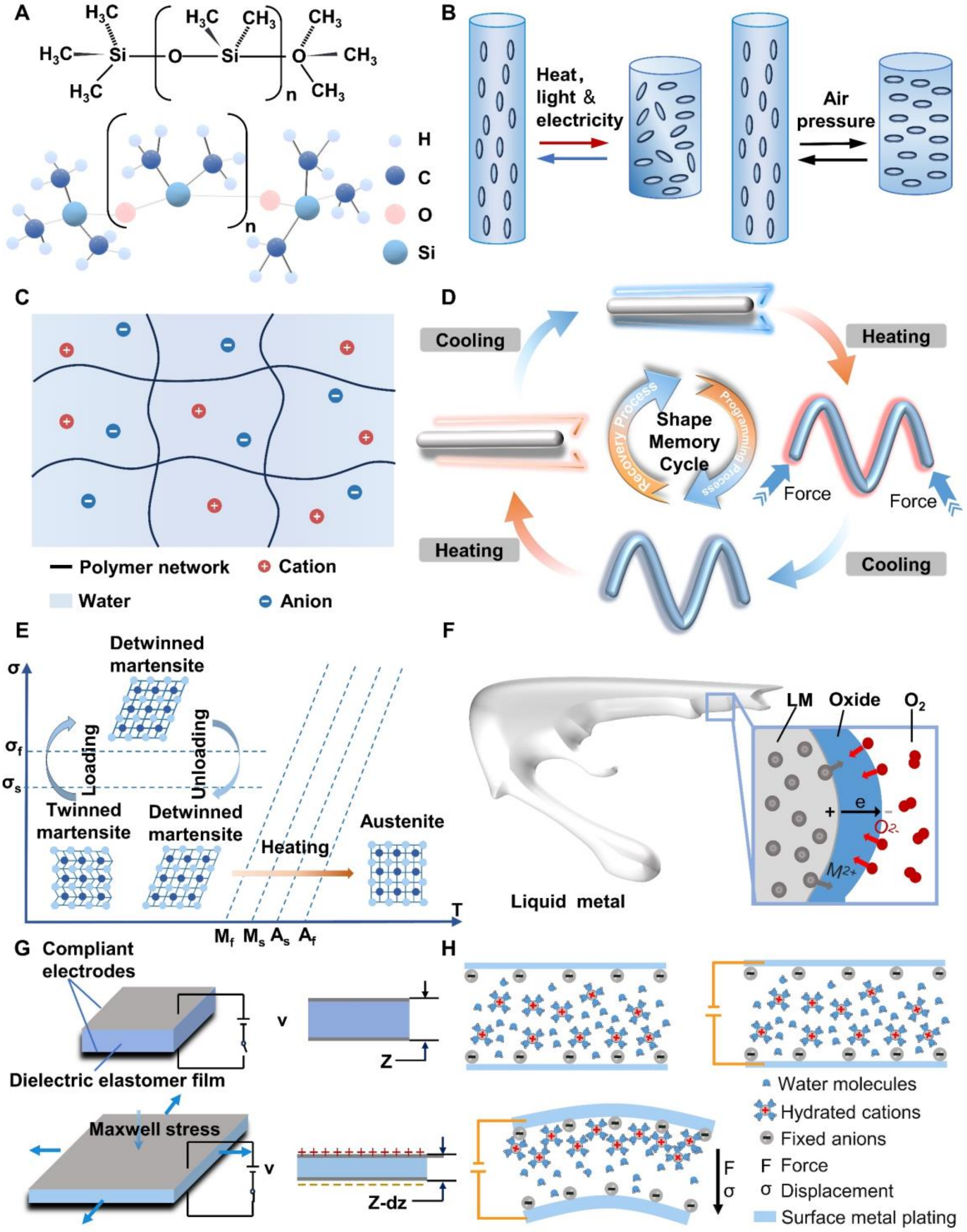

**Figure 3. Principles of the various soft materials.** (A) PDMS molecular structure, modified from Lu et al., 2023[106]. (B) LCEs, modified from Ma et al., 2025[107]. (C) Hydrogel, modified from Lee et al., 2020[108]. (D) SMP, modified from Xia et al., 2021[109]. (E) SMA,

modified from Chu et al., 2012[110]. (F) LM, modified from Vaillant et al., 2024[111]. (G) DE, modified from Shigemune et al., 2018[112]. (H) IPMC, modified from Chen et al., 2008[113].

*3.1.1 Elastomers*

Elastomers are fundamental to soft robots, offering high elasticity, recoverability, and adaptability. Common elastomers used in biomimetic underwater robots include silicone, polydimethylsiloxane (PDMS), polyurethane (PU), polyacrylate (PA), styrene-ethylene-butylene-styrene (SEBS), and liquid crystal elastomers (LCEs).

Silicone is widely employed for its flexibility, biocompatibility, and water resistance. For example, it has been used to seal electronics in a deep-sea lionfish-inspired robot[64] and to enhance adaptive attachment in a biomimetic remora suction disc[76]. PDMS offers high optical transparency (>92% transmission) and compatibility with micro/nanofabrication, making it ideal for soft sensors and actuators (**Figure 3A**)[106]. For instance, it has served as an encapsulation layer and functional coating for transparent dielectric elastomer actuators in the design of a jellyfish robot[114]. PU provides excellent wear resistance but has limited chemical stability. PA offers high stretchability and viscoelasticity, making it useful for dielectric layers in flexible devices. SEBS features high ultraviolet resistance and elongation at break exceeding 1000%, making it well-suited for flexible actuators.

LCEs are thermoplastic elastomers that combine entropy elasticity and self-organization, with liquid crystal molecules acting as physical cross-links[115,116]. By adjusting the alignment of the liquid crystal units and the polymer network's elasticity, LCEs exhibit both flexibility and self-organizing properties, enabling significant deformation in response to external stimuli (Figure 3B)[107]. These materials show remarkable features such as large deformation, shape memory, and self-healing, making them promising for soft robots. LCEs can respond to light, heat, photothermal, and photochemical stimuli. They are especially sensitive to temperature, transitioning reversibly from an ordered anisotropic phase (stretching) to a disordered isotropic phase (shrinking) near their phase transition temperature. The first LCEs, synthesized by Finkelmann et al.[117] in 1981, exhibited a 26% shrinkage strain upon heating. Utilizing LCE to fabricate a bionic fish-tail structure, enabling the bio-inspired soft robotic fish to swim on the water surface[118].

*3.1.2 Hydrogels*

Hydrogels are crosslinked polymer network materials containing 80-90% water[119], capable of reversibly adsorbing and releasing water via hydrogen bonds (Figure 3C)[108]. This unique liquid-solid coupling property grants them high stretchability, toughness, self-healing ability, controllable biodegradability, and compatibility with 3D printing. An in-situ free liquid 3D printing technique based on yield-stress hydrogels has been developed to print complex functional components, such as high-aspect-ratio sensor tentacles and gripper hooks, directly onto assembled soft robots, enabling functional upgrades and in-situ repairs[120]. The hydrogel's properties ensure structural stability during printing, while it can be quickly removed with water post-printing, reducing material waste by over 90%. Hydrogels also exhibit exceptional transparency and responsiveness to external stimuli such as heat, pH, light, chemicals, and electricity, enabling reversible shape changes for smart actuation. A bio-inspired robotic fish fabricated with hydrogel material achieves underwater camouflage through its transparent body[121]. Some hydrogels possess self-healing properties akin to biological tissues.

*3.1.3 Smart materials*

Smart materials are crucial for underwater soft robots due to their ability to respond to environmental stimuli, enabling adaptive changes in shape, color, and electrical or magnetic properties. They primarily include shape memory materials (SMMs), liquid metals, and electroactive polymers (EAPs).

SMMs, including shape memory polymers (SMPs) and shape memory alloys (SMAs), can return to a preset shape when triggered. SMPs can be temporarily deformed upon heating, fixed upon cooling, and restored when reheated (Figure 3D)[109]. SMP serves as the core artificial actuator in the fabrication of bistable swimming robots[122]. SMAs undergo a reversible phase transition between high-temperature austenite and low-temperature martensite phases (Figure 3E)[110], enabling precise, stimulus-responsive actuation. SMA springs are employed as artificial actuators to mimic the transverse muscles of the octopus arm[123].

Liquid metals, such as gallium-based alloys, offer low melting points, high conductivity, and excellent deformability[124]. Their oxide layer increases surface tension, allowing non-spherical morphologies (Figure 3F)[111]. Embedded in elastomeric channels, they create stretchable conductors that retain high conductivity while preserving elasticity. An artificial muscle using electrochemical control of liquid metal interfacial tension enables robotic fish propulsion at 15 cm/min by mimicking biological muscle contraction[125].

EAPs, which deform under an electric field, are classified as field-activated (dielectric elastomers, DEs) or ionically activated (ionic polymer-metal composites, IPMCs). DEs contract in thickness and expand in area under voltage application (Figure 3G)[112], achieving strains up to 380% for high-performance actuation. A soft eel robot leveraging a liquid-electrode interface demonstrated efficient propulsion at 1.9 mm/s via traveling wave actuation, overcoming traditional DE actuator constraints[126]. IPMCs, composed of ion-exchange membranes and metal electrodes, deform at low voltages due to asymmetric ionic migration, generating internal stress imbalances that induce bending (Figure 3H)[113]. Their ability to function as both actuators and sensors makes them ideal for underwater robots. IPMC materials actuate the pectoral fins of a robotic manta ray[127]. Independent control of multiple IPMC beams, integrated with a flexible PDMS membrane, enables 3D undulatory propulsion for untethered underwater swimming.

**Table 1.** Comparison of various material properties.

| Material category | Material name | Characteristics | | | | | | |
|---|---|---|---|---|---|---|---|---|
| | | Quantitative perspective | | | | | Qualitative perspective | Refs. |
| | | Density [g/cm$^3$] | Modulus [MPa] | Actuation strain [%] | Power density (kW/m$^3$) | Response time (s) | | |
| Elastomers | PDMS | 0.97-1.1 | 1-3[128] | Medium (1-10$^3$) | 10-100 | Slow | -Excellent optical transparency<br>-Outstanding biocompatibility<br>-Low young's modulus enables exceptional structural flexibility<br>-Facilitates bonding with diverse substrates at ambient temperature | [114,129] |
| | LCEs[130] | About 1.2[131] | 0.1-3 | Large (10-50) | 0.01-10 | Medium | -Stimuli-responsive large deformation<br>-Programmable shape memory with | [118] |
| | | | | | | | high recovery rate<br>-Intrinsic self-healing capability | |
| Hydrogels | Hydrogels | Highly variable depending on water content | | | | | -Fluid–solid coupling<br>-Ultrahigh stretchability with dynamic bond-reinforced toughness<br>-Cornea-like optical transparency<br>-Stimuli-triggered reversible shape morphing with programmable trajectories | [121,132] |
| Intelligent materials | SMA[130] | 5.1-6.7[133] | 28-75 × $10^3$ | Medium (4-8) | $10^3$-$10^5$ | Variable | -High power density<br>-Elevated mechanical stress<br>-Reversible deformation with shape memory effect<br>-No noise | [123,134,135] |
| | SMP | 0.9-1.3[136] | Varies widely by composition | Very large ($>10^3$)[137] | ($10^{-2}$ × $10^3$)[138] | Slow | -Excellent elastic deformability<br>-Low density low cost<br>-Biocompatibility<br>-Manufacturing-friendly<br>-Adjustable transition temperature | [122] |
| | DE[130] | 1-2 | 0.1-3 | Large (1-$10^3$) | 100-3500 | Fast | -High output stress<br>-Requires high activation electric fields<br>-Fast response time | [139,140] |
| | IPMC | 1.0-2.25[141] | (0.1-1 Hz) Small<br>(5-20 Hz) Large[142] | Large (>40) | 0.01-1 | Medium/ Fast | -Low-voltage-induced large deformation<br>-Low operating voltage (1-5V)<br>-High-frequency operation (100Hz)<br>-Aqueous environment compatibility | [127,143] |

### 3.2 Actuation

Actuators are fundamental to soft robots, enabling continuous, multi-degree-of-freedom deformations such as bending, stretching, and twisting. By utilizing low-elastic-modulus materials that respond to external stimuli, soft actuators overcome the constraints of traditional rigid motor-link systems. These actuation mechanisms exploit material properties like superelasticity and nonlinear strain responses triggered by pressure, electromagnetic, or thermal fields. Common approaches include fluidic, cable-driven, SMMs, EAP, magnetostrictive materials, chemical actuation, biomaterial-based actuation, and hybrid actuation (**Table 2**).

*3.2.1 Fluidic actuation*

Fluidic actuation integrates pneumatic or hydraulic channels within elastomers to replicate soft-bodied animal movements. Pneumatic and hydraulic actuators (PA and HA) use braided layers around elastic tubes for controlled stretching and contraction, while strain-limiting elastomer layers enable precise bending or twisting. For examples, Wang et al.[144] developed a pneumatic-driven biomimetic remora sucker, achieving strong underwater attachment and rapid curling detachment within 1 second. Xie et al.[145] created a pneumatic-driven bioinspired octopus arm that enables adaptive grasping using a bending motion (**Figure 4A**). Marchese et al.[146] developed a soft robotic fish with integrated micro-pump actuators for efficient fluidic propulsion (Figure 4B).

*3.2.2 Cable-driven actuation*

Inspired by the human tendon system, cable-driven actuation generates bending and elongation by tensioning flexible cables anchored at fixed points. These systems use motors, transmission structures, and flexible cables, enabling multimodal deformations. Depending on cable arrangement and action, common cable actuators are classified into longitudinal and lateral stretch actuators. Longitudinal stretch actuators generate bending movements when the cable is pulled, while lateral stretch actuators produce elongation. By controlling the motor's rotation, the tension and length of the cable can be adjusted, enabling a variety of movements such as bending, twisting, or contraction in the soft structure. For instance, a bioinspired octopus arm robot with embedded cable actuators replicates complex movements (Figure 4C), demonstrating high maneuverability in underwater environments[147,148].

*3.2.3 SMM actuation*

SMM actuators use the material's intrinsic stimulus-response properties to achieve motion and deformation. SMA actuators work through the thermomechanical coupling of shape memory alloys, where heating and cooling cause stretching and contracting, generating the driving force for flexible robots[149]. SMPs, developed in the mid-1980s[150], are considered smart polymers that respond to external stimuli (heat[151], light[152], magnetism[153], electricity[154], humidity[155]), exhibiting dynamic shape memory characteristics. SMP actuators leverage the shape memory effect triggered by specific stimuli, with thermal activation being the most common. Huang et al.[156] designed a frog robot using SMA for untethered swimming. Patterson et al.[134] developed PATRICK, a brittle star-like soft robot with SMA-actuated silicone legs (Figure 4D). Chen et al.[122] introduced a soft swimming robot driven by SMP actuators and bistable elements, utilizing temperature-responsive actuation for preprogrammed tasks without onboard electronics (Figure 4E).

*3.2.4 Chemical actuation*

Chemical actuation relies on endogenous chemical reactions, such as catalytic decomposition or redox reactions, to achieve autonomous propulsion without requiring external energy fields. These robots convert the energy produced by chemical reactions into the mechanical energy needed for movement. For example, Wehner et al.[157] developed the Octobot, a 3D-printed soft octopus-like robot powered by chemically generated gas. A microfluidic system regulates hydrogen peroxide flow, which decomposes catalytically to produce gas, inflating internal channels and driving arm movement for underwater propulsion. Liu et al.[158] developed a jellyfish-shaped soft robot powered by aqueous $ZnBr_2/ZnI_2$ redox flow batteries (Figure 4F). The liquid electrolyte, comprising 90% of the robot's mass, circulates through 3D-printed channels during motor-driven bell contractions, enabling controlled $Zn^{2+}$/halide redox reactions. This system generates a power density of up to 150 mW $cm^{-2}$, driving silicone tendons to achieve sustained underwater propulsion at 2 cm/s for 1.5 hours.

*3.2.5 EAP actuation*

EAP materials can change their internal structure under an electric field, producing various mechanical responses such as stretching, bending, tightening, or expanding, making them suitable for soft robot actuators. The two main types of EAP actuators, DE and IPMC, have numerous applications[159]. Li et al.[160] designed a artificial muscle actuator based on the biomechanics of rays, using pre-stretched dual-layer DE films to drive silicone fish fins for undulatory propulsion via periodic electric field excitation (Figure 4G). The IPMC-driven manta ray robot developed by Chen et al.[161] uses four IPMC fins to drive the chest fins, propelling a 180 mm wingspan body at 4.2 mm/s (Figure 4H). Shen et al.[143] designed a underwater soft robot with switchable swimming modes using IPMC artificial muscles. The robot's two soft fins are driven by six embedded IPMC actuators connected to silicone membranes, utilizing the multiple shape memory effects of IPMC materials to alter swimming modes and improve maneuverability.

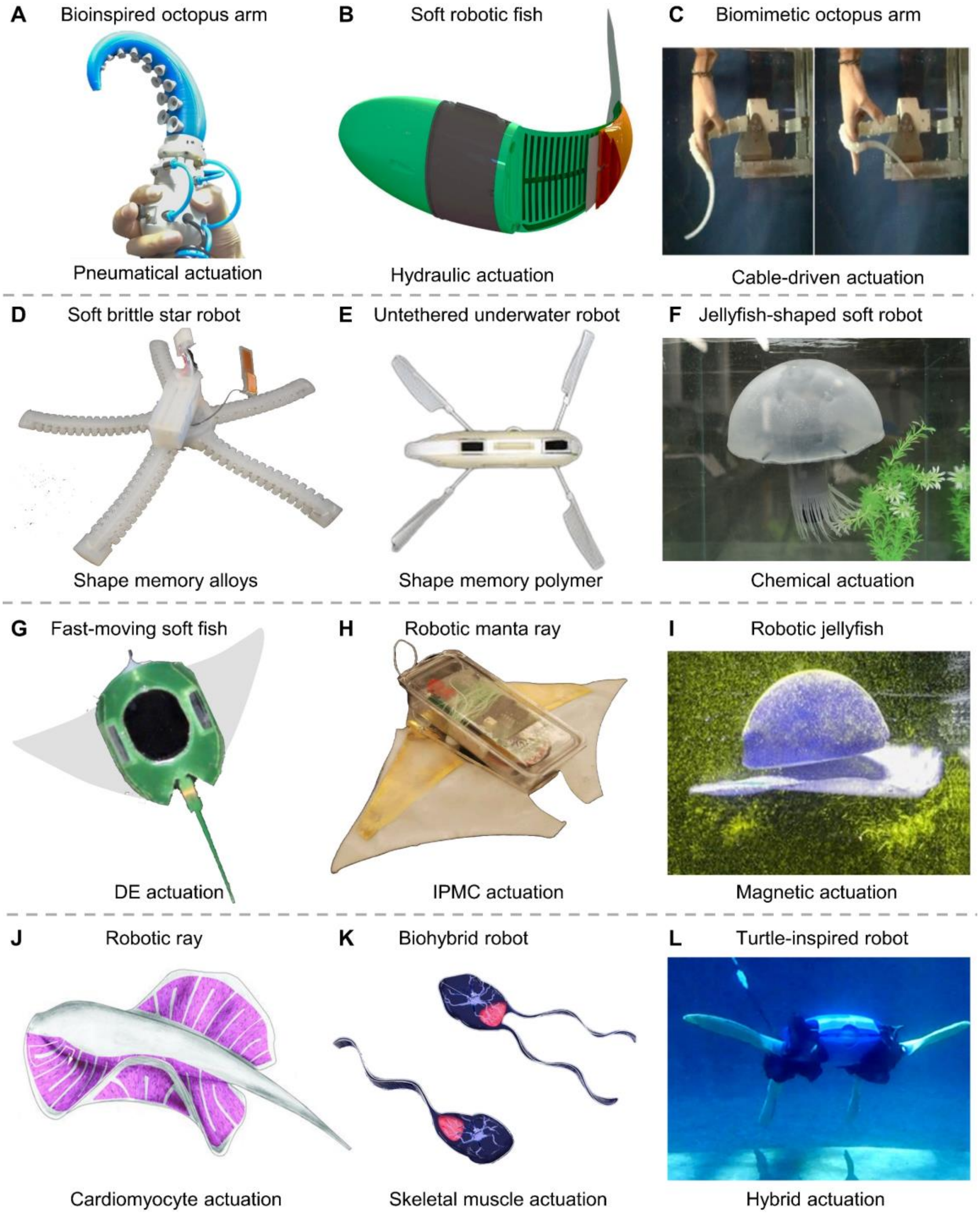

**Figure 4. Bioinspired underwater soft robots with representative actuation mechanisms.**

(A) Pneumatical actuation. Reproduced with permission.[145] Copyright 2020, Mary Ann Liebert, Inc. (B) Hydraulic actuation. Reproduced with permission.[146] Copyright 2014, Mary Ann Liebert, Inc. (C) Cable-driven actuation. Reproduced with permission.[147,148] Copyright 2017, IEEE. (D) SMA actuation. Reproduced with permission.[134] Copyright 2020, IEEE. (E) SMP actuation. Reproduced under the terms of a PNAS license.[122] Copyright 2018, National Academy of Sciences. (F) Chemical actuation. Reproduced with permission.[158] Copyright 2024, AAAS. (G) DE actuation. Reproduced with permission.[160] Copyright 2017, AAAS. (H) IPMC actuation. Reproduced with permission.[161] Copyright 2012, Taylor&Francis. (I) Magnetic actuation. Reproduced with permission.[162] Copyright 2022, Mary Ann Liebert, Inc. (J) Cardiomyocytes actuation. Reproduced with permission.[163] Copyright 2016, AAAS. (K) Skeletal muscle actuation. Reproduced under the terms of a CC BY-NC-ND license.[164] Copyright 2019, National Academy of Sciences. (L) Hybrid actuation. Reproduced with permission.[165] Copyright 2022, Springer Nature.

#### *3.2.6 Magnetic actuation*

Magnetic actuation mechanism is based on the dynamic coupling of magnetic fields and soft materials. Using permanent magnets, electromagnets, or Helmholtz coils, a magnetic field control system enables multidimensional motion by switching between uniform, gradient, or rotational fields. The arrangement of magnetic particles in the hydrogel/elastomer matrix allows programmable magnetic domains. In a gradient field, the robot moves directionally using magnetic gradient forces; in a rotating field, magnetic torque causes deformation, enabling rolling, contraction, or helical motion. This contactless driving approach overcomes traditional mechanical limits, offering precise control for applications like minimally invasive surgery and targeted drug delivery. Guo et al.[166] developed a liquid metal-based electromagnetic actuator, applying Lorentz forces to drive jellyfish and fish tail robots. Similarly, Ye et al.[162] designed the LM-Jelly robot, driven by alternating current to generate Lorentz forces, causing the membrane to contract and expand like a moon jellyfish (Figure 4I).

#### *3.2.7 Biomaterial actuation*

Biologically active materials, such as cardiomyocytes and skeletal muscle cells, offer biocompatible and efficient actuation for soft robots. These cells contract directionally in response to electrophysiological or chemical stimuli and can be integrated into flexible substrates or 3D scaffolds to form biohybrid actuators. Cardiomyocytes provide spontaneous rhythmic contractions for continuous microrobot motion, while skeletal muscle cells enable precise force control (up to hundreds of micronewtons per cell) through neural regulation. Tissue engineering techniques, including in vitro muscle sheets and 3D bundles, enhance deformation and force output, improving biomimetic motion efficiency. Park et al.[163] fabricated a light-driven biohybrid soft robotic ray, powered by optogenetically engineered cardiomyocytes, achieving 3.2 mm/s swimming speed with independent fin control (Figure 4J). Aydin et al.[164] created a neuromuscular-driven swimming robot that achieves autonomous locomotion in low Reynolds number environments by using optogenetic neurons to control rhythmic muscle contractions (Figure 4K). By optimizing a dual-tail structure through computational modeling, the system demonstrates precise neuro-muscular control of robotic motion.

Soft robotic actuation has advanced through diverse mechanisms, each offering unique advantages in energy efficiency, adaptability, and biomimetic motion. Fluidic, cable-driven, SMM, and EAP actuators enable versatile deformation control, while chemical and magnetic actuation facilitate untethered operation. Biomaterial actuation, leveraging living tissues, represents a promising frontier for biohybrid systems. Many advanced robots integrate multiple actuation strategies to enhance performance. For example, a sea turtle robot combines pneumatical actuation with thermally activated variable-stiffness materials, achieving efficient propulsion and adaptable limb rigidity (Figure 4L)[165]. As research progresses, hybrid actuation approaches will continue to drive innovation, unlocking new capabilities in underwater soft robotics.

**Table 2.** Advantages and limitations of the actuation methods.

| Actuation | | Advantages | Main limitations | Refs |
|---|---|---|---|---|
| **Fluidic actuation** | PAM | -Light air quality, widely available<br>-Non-polluting<br>-Simple structure | -Dependent on external air compressors and electromagnetic valves<br>-Large volume | [167,168] |
| | FEA | -Highly deformable | -Requires complex air | [146,169] |
| | | -Strong adaptability<br>-Low power consumption | compressors or hydraulic pumps<br>-High requirements for sealing | |
| **Cable-driven actuation** | | -Can achieve both bending and elongation depending on the design<br>-Proven effectiveness | -Complex and bulky systems<br>-Difficulty in precisely controlling the tension of each cable | [170,171] |
| **Shape memory material actuation** | SMA | -High power density and high stress | -Small range of motion, significant hysteresis, and creep | [172,173] |
| | SMP | -High elastic deformation and ease of manufacturing | -Low mechanical strength<br>-Low recovery stress (1-3 MPa) | [174,175] |
| **Chemical actuation** | | -Powerful power output<br>-No external energy input<br>-Autonomous movement, enhancing mobility and flexibility | -Reaction control complexity<br>-Limited runtime due to the finite amount of fuel carried | [176] |
| **Electroactive polymer actuation** | DE | -Large output stress | -Requires high excitation electric field | [177,178] |
| | IPMC | -Low driving voltage<br>-large deformation | -Generally small electromotive stress | [179,180] |
| **Magnetic actuation** | | -Untethered operation<br>-Ease of control<br>-Precise magnetic manipulation | -Tracking difficult<br>-Complex real-time field adjustments | [162], [132] |
| **Biomaterial actuation** | | -Safety and flexibility<br>-High propulsion efficiency<br>-Untethered | -Difficult in vitro culture<br>-Insufficient force output<br>-Dependence on external conditions | [164], [181] |

## 3.3 Fabrication

Advancements in bioinspired soft robot fabrication leverage cutting-edge techniques such as shape deposition modeling (SDM), 3D printing, soft lithography, mold casting, smart composite manufacturing, and biomanufacturing. These methods enhance precision, design flexibility, and functional integration.

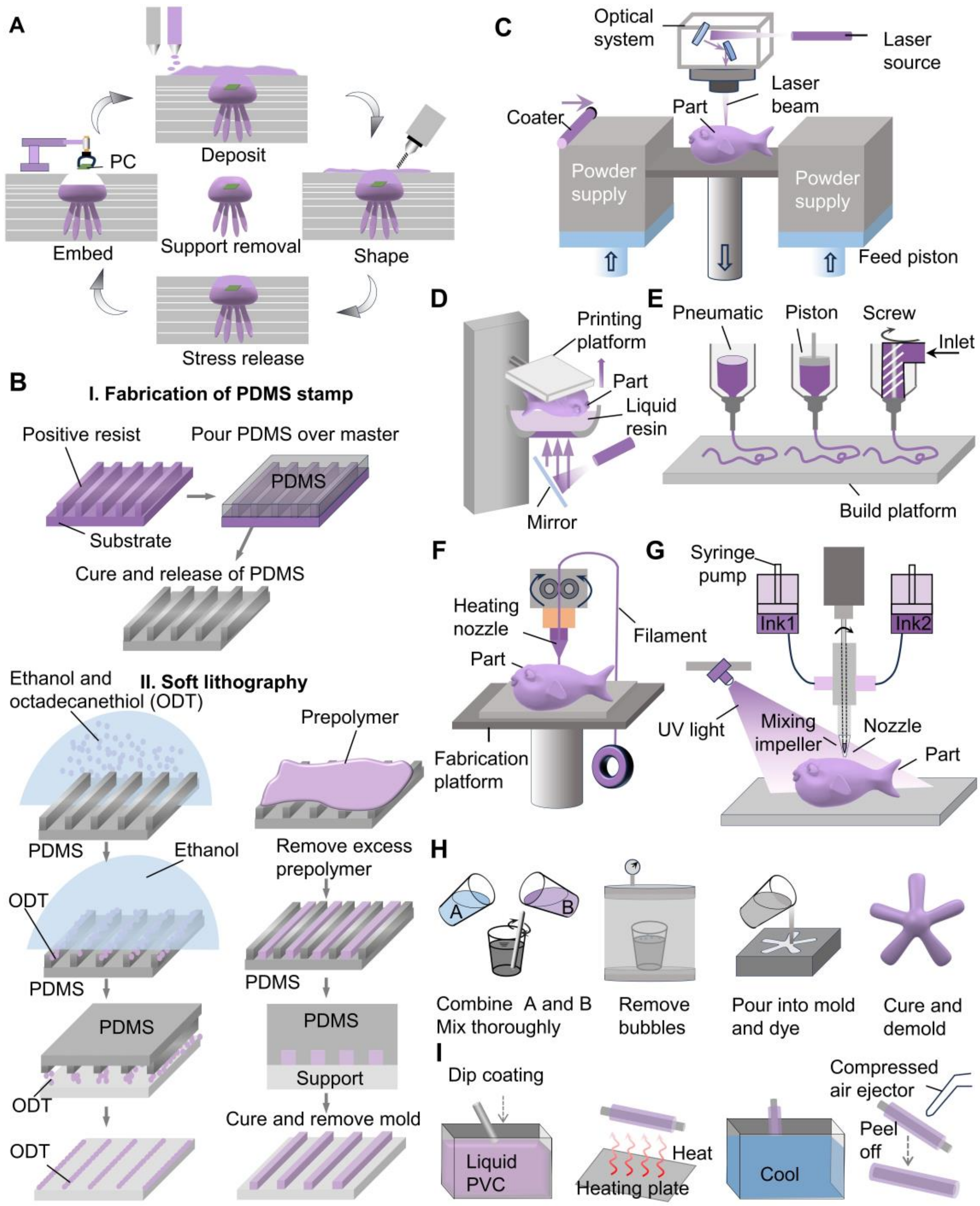

**Figure 5. Fabrication techniques for bioinspired underwater soft robots.** (A) Shape Deposition Manufacturing (SDM). (B) Soft lithography. (C) Selective Laser Sintering (SLS). (D) Stereolithography (SLA). (E) Direct Ink Writing (DIW). (F) Fused Deposition Writing (FDW). (G) Drop-on-Demand (DOD) printing. (H) Mold casting. (I) Dip coating.

### *3.3.1 Shape Deposition Manufacturing (SDM)*

Shape Deposition Manufacturing (SDM) is a rapid prototyping method that combines additive and subtractive techniques, enabling the layer-by-layer fabrication of complex 3D structures. Through alternating steps of material deposition, precision machining, and component embedding, SDM allows for the seamless integration of multiple materials, incorporating sensors and actuators directly into the structure. This iterative deposition-forming-machining cycle overcomes the spatial constraints of traditional manufacturing (**Figure 5A**). Since its introduction by Merz et al.[182] in 1994, SDM has been widely applied in soft robots. Kim et al.[183] used SDM to develop Stickybot, a biomimetic wall-climbing robot with body and limbs made from polyurethane materials of varying hardness.

*3.3.2 Soft lithography*

Soft lithography enables the repetitive transfer of complex microstructures by creating precise elastic molds. Unlike traditional photolithography, which involves steps like pattern design and exposure, soft lithography adds the process of creating a master mold from the substrate and then making the elastic mold (Figure 5B). Wehner et al.[157] used this technique to fabricate an octopus robot, integrating microfluidic channels into its arms via 3D printing. The advantage of soft lithography lies in its ability to produce intricate underwater soft robots at scale rapidly. Feng et al.[184] designed an underwater soft eel robot by fabricating a fiber-reinforced bidirectional bending fluidic elastomer actuator array using soft lithography. The core design employs a dual-chamber configuration (Ecoflex 00-30 silicone) integrated with a fiber winding reinforcement layer, achieving bidirectional bending (maximum bending angle of 120°) through hydraulic actuation.

*3.3.3 3D printing*

3D printing technology, with its layer-by-layer construction, has become a popular method for creating complex multiscale structures in soft robots and silicone molds[185,186]. This approach allows direct manufacturing from digital models, overcoming the limitations of traditional techniques on design flexibility. Common methods include selective laser sintering (SLS), stereolithography (SLA), direct ink writing (DIW), fused deposition modeling (FDM), and drop-on-demand (DOD).

SLS selectively fuses powdered material with a laser, creating intricate internal structures without support materials, though surface roughness often requires post-processing (Figure 5C). SLA uses photosensitive resins and a laser or projector to selectively harden the material, producing high-resolution, smooth surfaces, perfect for soft components with fine structures (Figure 5D). Peele et al.[187] used digital mask projection stereolithography to fabricate a multi-degree-of-freedom soft pneumatic actuator with a complex internal structure. DIW extrudes liquid “ink” materials along a predefined path, making it ideal for materials like hydrogels and silicones (Figure 5E). FDM uses molten thermoplastic materials to build objects layer by layer, suitable for various thermoplastics, including thermoplastic elastomers (Figure 5F). Yap et al.[188] introduced a low-cost method to fabricate high-force soft pneumatic actuators using FDM 3D printing. DOD creates objects by jetting microscopic droplets that solidify rapidly, enabling high-precision, multi-material printing, such as dual-ink systems. Solidification can be induced by various methods, including UV curing and temperature control (Figure 5G).

*3.3.4 Casting*

Soft robots rely heavily on materials like silicone rubber, making casting a widely used fabrication method. Gravity casting is the most common approach, where a 3D-printed mold is filled with a vacuum-degassed silicone mixture in a fixed ratio, cured, and then demolded (Figure 5H). However, demolding becomes challenging for complex shapes and internal cavities. Rotational casting addresses this by rotating the mold around an axis, allowing liquid silicone to coat the inner surface and gradually cure, forming intricate cavities without a core. However, it struggles with precise wall thickness control and is mainly suited for thin-walled structures. Similarly, dip coating is used for small or thin-shell structures by dipping a mold into liquid polymer, rapidly curing it, and then removing the shell (Figure 5I). While soft robots’ flexibility aids demolding, highly complex geometries remain challenging.

To overcome these limitations, split molds[189], flexible cores (foam cores)[190], and removable molds (soluble or fusible cores)[191] have been adopted from traditional casting methods to simplify demolding. The lost-wax method is particularly effective for creating intricate internal structures. Here, low-melting-point wax forms internal channels, which are embedded in silicone and later melted away after curing. This technique not only enables more complex cavities but also enhances structural strength due to its monolithic design. Katzschmann et al.[192] used lost-wax casting to fabricate the flexible caudal fin of a soft robotic fish, eliminating seam weaknesses and enabling precise internal channel geometry. This ensured high durability under repeated hydraulic actuation while supporting the continuous, compliant bending needed for bioinspired undulatory propulsion. Additionally, a stepwise casting process can be used to create complex structures by casting multiple simple structures, which are then bonded or impregnated with uncured material to seal them together.

*3.3.5 Bio-hybrid fabrication*

By combining bioactive materials (hydrogels and cell-encapsulated structures) with synthetic materials (flexible polymers and conductive fibers) across multiple scales, hybrid manufacturing techniques like 3D bioprinting and microfluidic assembly are used to create bioinspired tissue-machine interfaces. This approach enables soft robots to exhibit life-like characteristics, including self-healing, environmental adaptability (deformation in response to temperature or ion concentration), and autonomous energy conversion (light-driven or chemical energy-driven actuation), thereby significantly enhancing their ability to interact with, adapt to, and robustly operate in complex aquatic environments. Nawroth et al.[193] developed a biohybrid jellyfish robot by integrating neonatal rat cardiomyocytes with synthetic materials. Using microcontact printing, they patterned fibronectin on a PDMS elastic substrate to guide the radial alignment of cardiomyocytes into functional muscle tissue. Electrical stimulation synchronized cell contractions, recreating the undulatory motion of jellyfish. The robot replicated jellyfish swimming (0.4-0.7 body lengths per second) and feeding hydrodynamics, achieving propulsion via an alternating contraction-elastic recoil mechanism with efficiency comparable to real jellyfish larvae.

**3.4 Sensing**

The autonomy and environmental adaptability of bioinspired underwater soft robots rely on the performance of their sensing systems. Given the complex hydrodynamics and unpredictable biological interactions in aquatic environments, real-time sensing feedback is essential for closed-loop control, precise motion regulation, and intelligent environmental interaction. Unlike traditional rigid robots that use encoders and force sensors, soft robots require sensing technologies that accommodate their continuous deformation and compliant materials. These sensors must exhibit high stretchability, low modulus, and resistance to water pressure and corrosion while preserving the robot’s streamlined biomimetic design and motion freedom. The advancement of sensing technology in soft robots encompasses various sensor types, with core mechanisms including resistive, capacitive, optical, magnetic, inductive, and triboelectric sensing. The integration of flexible and stretchable sensors enables soft robots to perceive their own shape (proprioception) and detect external stimuli (exteroception).

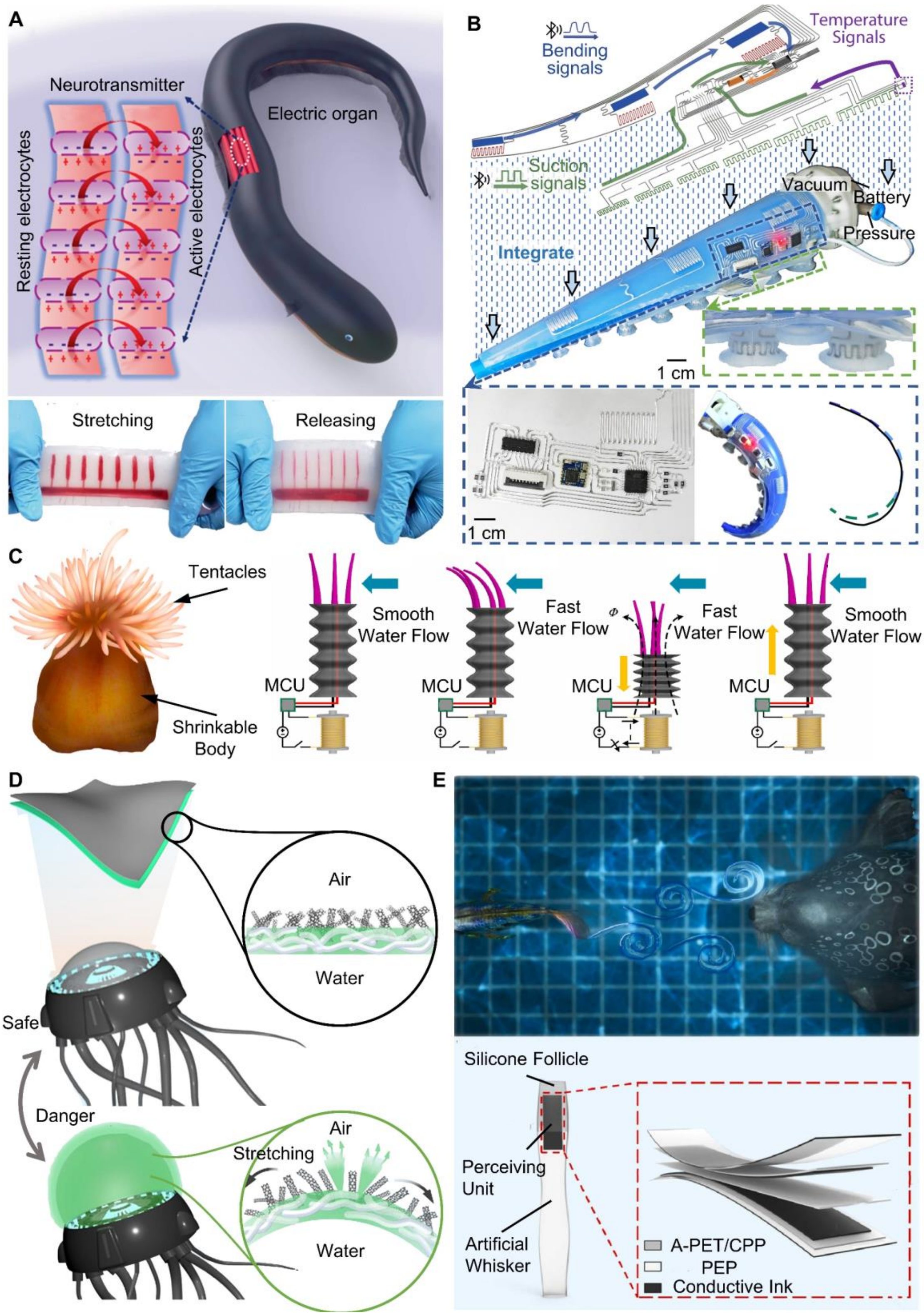

**Figure 6. Bioinspired underwater soft sensors and robots.** (A) Stretchable nanogenerator inspired by the electric eel. Reproduced under the terms of a CC-BY license.[194] Copyright 2019, Springer Nature. (B) Electroactive soft octopus arm integrated with liquid metal-based

flexible sensors. Reproduced with permission.[77] Copyright 2023, AAAS. (C) Tentacle sensors inspired by sea anemones. Reproduced with permission.[195] Copyright 2021, Elsevier. (D) Soft, luminescent and morphing skin inspired by jellyfish. Reproduced with permission.[196] Copyright 2024, American Chemical Society. (E) Triboelectric nanogenerator (TENG)-based underwater whisker sensor. Reproduced with permission.[197] Copyright 2022, Elsevier.

For proprioception, research has focused on actuators, McKibben muscles, and soft continuum robots. Helps and Rossiter[198] developed ionically driven soft bending and elongation actuators capable of self-sensing by monitoring resistance changes in the driving fluid. Zou et al.[194] replicated electric eel ion channels to design a bioinspired stretchable nanogenerator (BSNG) for underwater self-powered sensing (**Figure 6A**). A PDMS-silicone bilayer mimics ion channel stress responses, generating electrical signals via triboelectrification and electrostatic induction. Attached to soft robots, the BSNG functions as electronic skin, enabling real-time shape monitoring. Inspired by the octopus nervous system, Xie et al.[77] developed an electroactive soft octopus arm integrating multimodal sensing via liquid metal-based flexible circuits (Figure 6B). Embedded stretch and expansion sensors reconstruct 3D bending profiles for proprioceptive sensing. Liquid metal channels within the suction cups detect contact pressure through resistance changes, sensing adhesion status. Additionally, built-in temperature sensors transmit thermal gradient signals via liquid metal circuits, enabling external environmental perception.

For exteroception, Wang et al.[195] mimicked sea anemones' tentacle sensing and body contraction to develop a soft robot integrating magneto-electric sensing tentacles and a magnetically controlled shrinkable body (Figure 6C). Flexible silver nanoparticle sensors detect water flow by generating microvolt-level signals (2.8-7.3 μV) in response to magnetic field changes. The robot's NdFeB/Ecoflex spiral body contracts up to 64% under an external magnetic field, mimicking anemone retraction. It autonomously monitors flow velocity and contracts when speeds exceed 1.1 m/s, integrating self-powered sensing with adaptive deformation for stable underwater operation.

Li et al.[196] replicated the bioluminescent and morphing skin of Atolla jellyfish to create a soft underwater robot with self-sensing, autonomous 3D deformation, and fluorescent feedback (Figure 6D). A carbon nanotube network/fluorescent elastomer composite film mimics strain-responsive skin, modulating resistance and fluorescence under pressure changes (7-fold fluorescence increase at 160% strain). This closed-loop system converts hydraulic pressure into electrical signals, providing real-time depth sensing and visual warnings. Inspired by seal vibrissae, Wang et al.[197] developed a triboelectric nanogenerator (TENG)-based whisker sensor for passive vortex detection (Figure 6E). Vortex-induced bending triggers triboelectric charge separation at the fluorinated ethylene propylene (FEP)-water interface, generating signals that infer target motion. An electrostatic shield minimizes signal loss (≤12% at 2 Hz), while resonance frequency tuning optimizes sensitivity. Power-free and real-time, the sensor enables LED-based motion indication and underwater target tracking, demonstrating bioinspired passive hydrodynamic sensing.

### 3.5 Modeling and Control

Bioinspired underwater soft robots possess infinite degrees of freedom and highly nonlinear characteristics, making it challenging to translate sensory data into precise control commands. As a result, soft robot control remains a significant research challenge and can be broadly categorized into open-loop and closed-loop approaches.

Open-loop control, based on model-driven analysis, is widely used for motion control. The finite element method (FEM), a typical approach, discretizes infinite degrees of freedom into a finite set of elements. Du et al.[199] optimized the dynamic model and open-loop control signals using gradient data obtained from a differentiable simulator, reducing simulation-to-reality discrepancies and improving controller performance in a starfish-inspired quadrupedal swimming robot.

Closed-loop control, relying on feedback mechanisms, offers greater disturbance resistance and higher precision than open-loop control. Patterson et al.[200] developed PATRICK, an untethered bioinspired crawling soft robot, demonstrating the feasibility of closed-loop motion planning. Closed-loop control is further divided into single-level and multi-level architectures, with the latter enhancing system robustness through task decoupling. Huang et al.[201] improved control dimensions in a bioinspired frog robot by integrating posture sensing with online simulation, enabling real-time strategy adjustments in dynamic environments. This “perception-decision-execution” closed-loop paradigm has proven effective for adaptive underwater robot control.

Controller design is typically structured into three levels. Low-level controllers drive actuators, mid-level controllers manage kinematic and dynamic control, and high-level controllers handle task planning (obstacle avoidance). Soft robot control strategies can be classified by control target (static vs. dynamic controllers) and modeling approach (model-based, model-free, and hybrid controllers) [146].

Model-based control has evolved from static to dynamic approaches. Static controllers, based on simplified models like constant curvature assumptions, work well for uniform soft robotic arms but struggle with transient dynamics. Dynamic controllers, such as model predictive control (MPC), are widely used in industrial applications but face accuracy and real-time adaptability challenges when dealing with nonlinear, underactuated soft robot[202].

These challenges have driven the development of data-driven control paradigms, where machine learning enables adaptive control without precise models. Inverse kinematics learning (IKL) and forward dynamics learning (FDL) improve motion prediction, while reinforcement learning (RL) allows robots to autonomously optimize strategies through environmental interactions. Gu et al.[203] integrated deep reinforcement learning into a deformation control framework, enabling rapid caudal fin adjustments for bioinspired swimming. Li et al.[204] applied deep reinforcement learning to a dielectric elastomer swimming robot, optimizing motion in complex flow conditions.

Soft robot control remains a complex and evolving field, requiring a balance of precision, adaptability, and real-time performance. While model-based approaches provide theoretical rigor, they struggle with soft robots' inherent nonlinearity and deformability. Data-driven methods, particularly reinforcement learning, offer promising solutions for autonomous adaptation and optimization. The integration of machine learning and real-time sensing is expected to further advance closed-loop control, enhancing the autonomy and efficiency of soft robots in underwater exploration and complex operational tasks.

### 3.6 Power

Biological organisms achieve autonomous movement and adaptability by efficiently converting chemical and light energy into mechanical energy through complex, multifunctional systems. In contrast, modern robots lack such efficient energy conversion mechanisms, limiting their autonomy and endurance. Energy storage and utilization remain fundamental research challenges in bioinspired underwater soft robots.

Batteries are the primary energy source for bioinspired underwater soft robots. Lee et al.[205] employed lithium-ion batteries to power thermoelectric devices, inducing phase-changing fluid transitions for buoyancy control, enabling a flatfish-like robot to achieve 3D locomotion (**Figure 7A**). To enhance energy density, Aubin et al.[206] developed a redox-based circulation system that integrates energy storage, hydraulic transport, and actuation within a single structure, extending operational duration to 36 hours (Figure 7B).

Beyond batteries, alternative energy sources offer enhanced adaptability and autonomy. Wang et al.[207] used a low-intensity magnetic field (0.5 mT) to drive a manta ray robot, where magnetic torque induced elastic deformation in a NdFeB/PDMS composite, enabling propulsion (Figure 7C). Lee et al.[181] used bioelectrical energy from cardiomyocytes to power a biohybrid fish robot, achieving self-sustained swimming at 15 mm/s for 108 days through autonomous cardiac contractions, while optogenetic control enabled external pacing (Figure 7D). Xia et al.[208] employed near-infrared (NIR) light for photothermal conversion in a starfish soft robot, where reduced graphene oxide nanosheets within a hydrogel absorbed light, triggering thermal expansion and contraction for movement (Figure 7E).

Hybrid energy approaches further enhance performance. Wang et al.[209] developed a sea anemone-like actuator powered by both NIR light and magnetic fields (Figure 7F). A carbon nanotube-impregnated poly(*N*-isopropylacrylamide) hydrogel gripper responded to light-induced heating, while a magnetic-driven stalk enabled drifting motion. Fan et al.[210] used compressed gas supplemented by lithium battery control to mimic frog-like swimming via pneumatic actuators (Figure 7G). Elhadad et al.[211] employed microbial fuel cells, where bacterial metabolism generated 135 $\mu W/cm^2$ of electricity, powering a water strider robot for water surface locomotion (Figure 7H). He et al.[212] developed Copebot, a copepod-like soft robot driven by oxygen-propane combustion. Explosive thrust enabled high-speed jumps (850 body lengths/s), with adjustable fins allowing complex maneuvers such as 120° turns and 360° rotations, mimicking copepod escape behavior (Figure 7I).

These diverse energy strategies highlight the potential of bioinspired solutions to enhance the autonomy, efficiency, and adaptability of underwater soft robots, paving the way for next-generation untethered robotic systems.

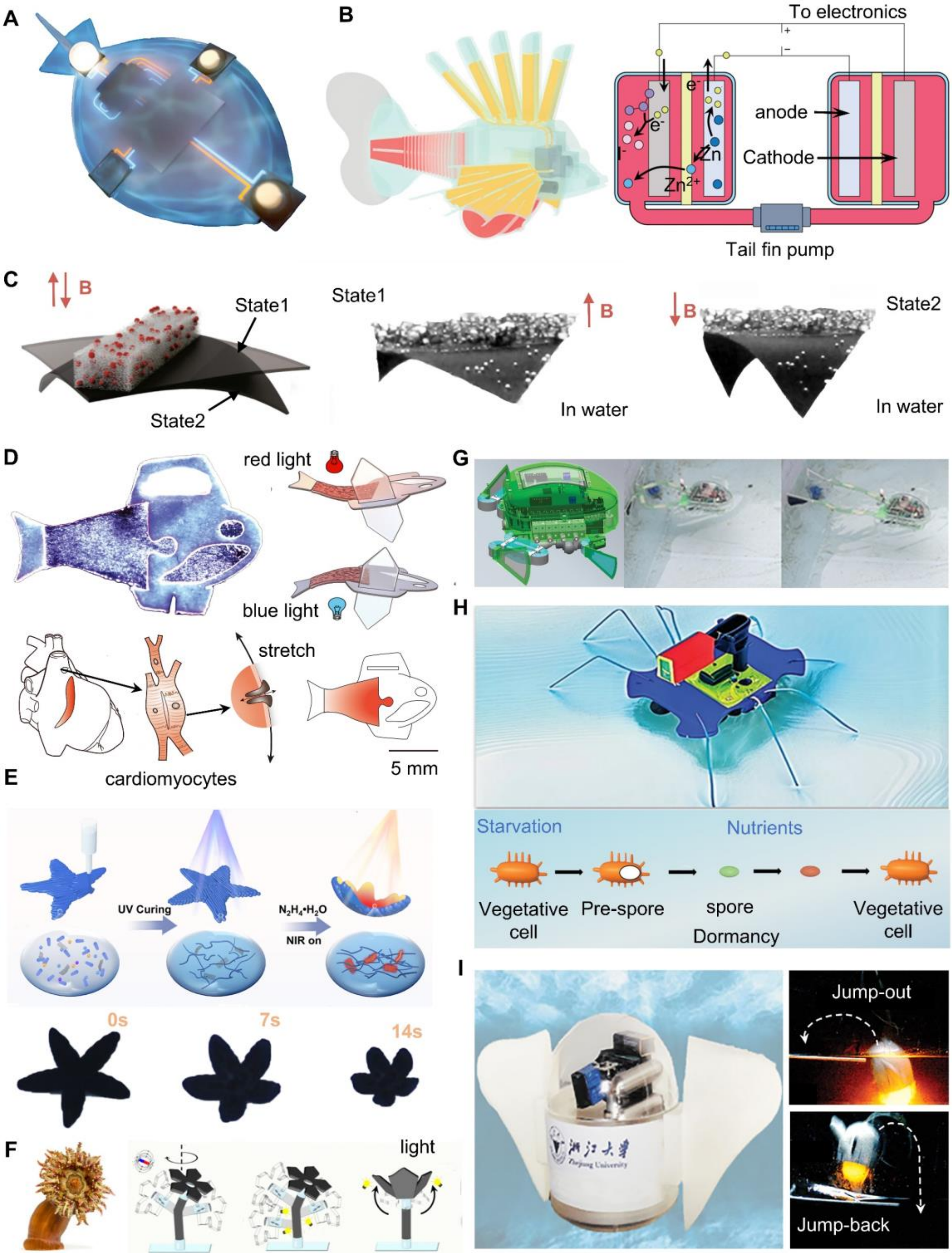

**Figure 7. Representative energy modes of underwater soft robots.** (A) Lithium-ion battery. Reproduced under the terms of a CC-BY license.[205] Copyright 2022, Wiley-VCH. (B) Redox-based circulation system. Reproduced with permission.[206] Copyright 2019, Springer Nature. (C) Magnetic field. Reproduced under the terms of a CC-BY license.[207] Copyright 2020, Springer Nature. (D) Bioelectrical energy. Reproduced with permission.[181] Copyright 2022, AAAS. (E) Near-infrared (NIR) light. Reproduced with permission.[208]

Copyright 2024, American Chemical Society. (F) Both NIR light and magnetic fields. Reproduced with permission.[209] Copyright 2023, American Chemical Society. (G) Compressed gas. Reproduced with permission.[210] Copyright 2020, Mary Ann Liebert, Inc. (H) Microbial fuel cells. Reproduced with permission.[211] Copyright 2024, Wiley-VCH. (I) Oxygen-propane combustion. Reproduced with permission.[212] Copyright 2024, Mary Ann Liebert, Inc.

## 4. Robotic insights into biological principles

Reverse engineering of biological systems has long been fundamental to bioinspired robots. Early research primarily focused on mechanically replicating biological features, such as morphology and movement, to address engineering challenges. However, with the rise of biomimetic soft robotics, the field has moved beyond superficial imitation to become a powerful experimental platform for biological scientific discovery. While biological systems inspire robotic design, robotics, in turn, provides quantitative frameworks to test biological hypotheses through physical modeling and controlled experiments—a methodology termed "robotics-inspired biology" by Lauder et al.[213] . This approach enables biomimetic soft robots to serve as experimental tools for studying material properties, morphological adaptations, locomotion mechanics, environmental interactions, sensory processing, and behavioral algorithms. Moreover, bioinspired robots offer a novel approach to paleontological research, enabling dynamic reconstructions of extinct species' locomotion and biomechanics. In parallel with the rise of "AI for Science", biomimetic robots are increasingly positioned as tools "for biological science"—offering controllable, repeatable, and measurable models to explore complex biological phenomena such as locomotion, sensing, and organism-environment interactions.

### 4.1 Robots as tools for biological research

In biomimetic underwater soft robots, reverse engineering of biological systems has not only advanced robotic performance but also provided deeper insights into biological principles. As discussed in Section 2, various biological features have been systematically extracted, forming the foundation for biomimetic robot design. The following examples demonstrate how these extracted features have been translated into soft robotic systems, which, in turn, have been used as experimental platforms to investigate the underlying biological mechanisms.

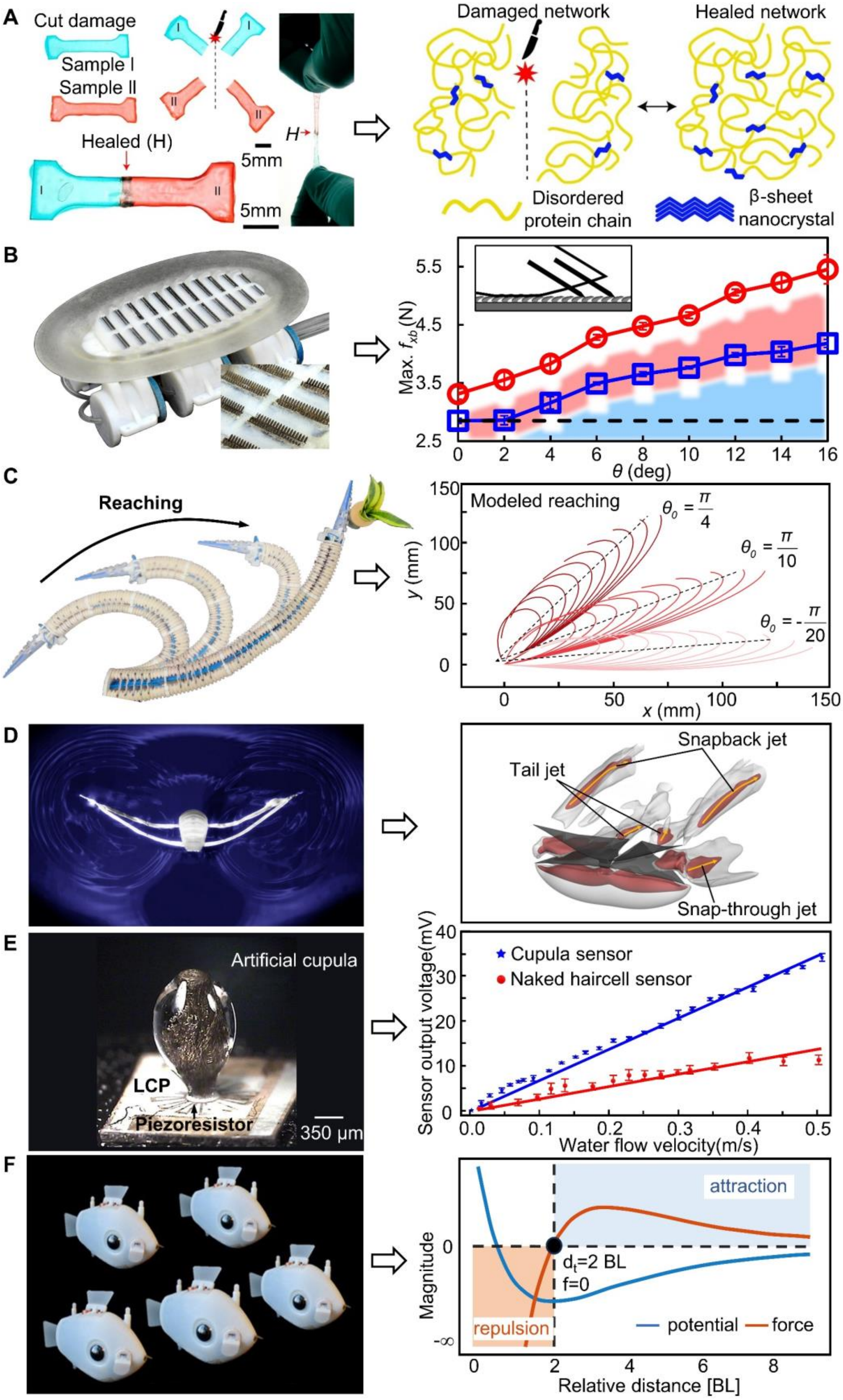

**Figure 8. Robots as experimental tools for unveiling biological mechanisms.** (A) Bioinspired squid sucker protein material reveals the ultrarapid self-healing mechanism of β-

sheet nanocrystal-reinforced supramolecular networks. Reproduced with permission.[74] Copyright 2020, Springer Nature. (B) Bioinspired remora adhesion disc reveals the friction enhancement mechanism of spinules. Reproduced with permission.[75] Copyright 2017, AAAS. (C) Bioinspired octopus arm reveals reaching locomotion mechanism. Reproduced with permission.[77] Copyright 2023, AAAS. (D) Bioinspired manta ray robot reveals vortex dipole-enhanced thrust and collision resilience via snapping-induced bifurcating jets. Reproduced with permission.[214] Copyright 2024, AAAS. (E) Bioinspired lateral-line sensor reveals cupula-mediated hydrodynamic signal amplification via drag-optimized β-sheet fibril networks. Reproduced under the terms of a CC- BY license.[215] Copyright 2016, Springer Nature. (F) Bioinspired Bluebot swarm reveals the density reconfiguration mechanism via adaptive Lennard-Jones-like potentials.[80] Copyright 2021, AAAS.

*4.1.1 Materials mechanism*

Inspired by the sucker protein material of squid, Abdon et al.[74] developed a synthetic self-healing protein material capable of repairing extreme damage, such as punctures and cuts, within one second at 50°C. Experimental studies of this material reveal the ultrarapid self-healing mechanism of β-sheet nanocrystal-reinforced supramolecular networks (**Figure 8A**). This bioinspired material outperforms natural self-healing proteins in strength and stability, making it suitable for underwater adaptive grippers. The protein film actuator enables precise grasping of fragile objects, rapid damage recovery, and controlled degradation under varying pH conditions. Similarly, Su et al.[4] designed a biomimetic disc with composite materials inspired by the remora's adhesive lip structure, embedding electrostatic flocking fibers into a soft silicone matrix. Their prototype demonstrated how vertically aligned collagen fibers significantly enhance adhesion performance.

*4.1.2 Morphology mechanism*

Based on the 3D scan morphological model of biological denticles, Wen et al.[5,216] fabricated 3D-printed flexible membranes with denticles, resulting in a biomimetic sharkskin. Using this biomimetic prototype, they explored the impact of different denticle arrangement patterns (interlocking, linear overlapping, and linear non-overlapping) and spacing parameters on fluid dynamics. They also discovered that the leading-edge vortex of the biomimetic sharkskin was stronger than that of a membrane without denticles, which could be a critical factor in enhancing propulsion. Wang et al.[75] replicated the macro- and microstructural morphology of the biological remora adhesive disc using multi-material 3D printing to develop a biomimetic sucker. Comparative experiments with and without spinules revealed the frictional force regulation mechanism of lamellae and spinules, explaining how remoras resist strong incoming water flow (Figure 8B). Similarly, Li et al.[76,217] investigated a biomimetic remora adhesion disc, uncovering the mechanism of redundant adhesion through independent lamellar compartments, which enables remoras to remain attached to marine organisms for extended periods.

*4.1.3 Kinematic mechanism*

In robotics kinematics research, Xie et al.[77] developed a pneumatically actuated octopus-inspired soft arm. Comprising a five-segment continuous soft-body structure with a distal gripper, the researchers established a biomimetic arm model and investigated its bending-elongation propagation mechanism through prototype experimentation (Figure 8C). Qing et al. developed a bioinspired manta ray soft robot with a monostable wing driven by pneumatic actuation and spontaneous snapback, uncovering a dual-jet thrust enhancement mechanism. CFD and PIV analyses reveal that the asymmetric square-wave motion generates bifurcated vortex pairs (2P wake) and dual high-momentum jets, enhancing thrust fourfold over sinusoidal motions (Figure 8D). Villanueva et al.[218] developed the biomimetic Robojelly, a soft jellyfish using shape memory alloys and silicone, replicating the moon jellyfish's bell contraction-relaxation dynamics. Their study showed that passive folding elasticity recovers 41% of energy, while a reverse Kármán vortex street enhances propulsion during contraction. Zhong et al.[219] used a cable-driven elastic actuator to simulate the tendon tension regulation strategy in biology, controlling the stiffness of a bioinspired tuna tail fin and revealing the kinematic mechanism of efficient propulsion. The variable stiffness mechanism reduces energy dissipation by optimizing the wake vortex structure (transition from attached leading-edge vortices to detached ring vortices).

*4.1.4 Sensor mechanism*

Mimicking the superficial neuromasts of blind cavefish, Kottapalli et al.[215] developed bioinspired hydrogel neuromast sensors utilizing a hyaluronic acid hydrogel cupula reinforced by an electrospun nanofibril scaffold to replicate the biological cupula's structure and mechanical properties. Wind/water tunnel tests showed that the hydrogel cupula enhanced drag force through increased surface area and hydrophilic-hydrodynamic interactions, while the nanofibril scaffold improved structural integrity and signal coupling. The sensor achieved 3.5-5 times higher sensitivity and 2 times improved resolution compared to bare hair cell sensors, demonstrating a biomimetic mechanotransduction mechanism where hydrogel viscoelasticity and fibril-assisted fluid-structure interaction enhance flow signal transmission, akin to biological neuromasts detecting boundary-layer-exceeding hydrodynamic stimuli (Figure 8E). Similarly, Peleshanko et al.[79] developed a biomimetic hair sensor inspired by the cavefish lateral line. They integrated a piezoresistive strain sensor within a polyethylene glycol hydrogel crown, replicating the gel crown’s morphology and function to significantly improve sensitivity and signal transmission efficiency. Asadnia et al.[220] created a biomimetic soft polymer nanosensor mimicking hair cell ciliary bundles. The design features gradient-height PDMS micropillars resembling ciliary morphology, interconnected by polyvinylidene fluoride (PVDF) nanofibers acting as artificial tip links. Fluid drag deflects the pillars, stretching the nanofibers to generate piezoelectric charges. A hyaluronic acid hydrogel cupula further enhances fluid-structure interaction, simulating biological mechanotransduction and converting fluid-induced ciliary deflection into electrical signals.

*4.1.5 Swarm mechanism*

Research on biomimetic underwater soft swarm robots is limited, with existing studies mainly focusing on biomimetic fish schools. Berlinger et al.[80] designed the Bluebot, a robotic unit featuring 3D visual perception (via dual-camera and LED-based positioning) and multi-fin coordinated propulsion, and assembled the Blueswarm, a fish-inspired robotic swarm. Utilizing local implicit coordination mechanisms (without centralized control or explicit communication), they demonstrated complex collective behaviors including synchronized flashing, potential field-driven dispersion/aggregation, dynamic circular formations, and search-capture tasks. Experiments validated the effectiveness of the Mirollo-Strogatz synchronization model, Lennard-Jones potential field modulation, and binary steering rules, revealing that dynamic global patterns can self-organize solely through local neighbor perception and simple behavioral feedback mechanisms (Figure 8F). Li et al.[105] used biomimetic fish schools to uncover the energy-saving mechanism of reverse Kármán vortex shedding synchronization (VPM). They showed that fish reduce swimming energy consumption by over 14% through tailbeat phase matching with the vortices of leading individuals, relying solely on proprioception and flow field feedback, without visual input.

**4.2 Robotics for paleontology**

Beyond studying living organisms, bioinspired robots provide a revolutionary tool for exploring the form and function of ancient organisms. Unlike the fragmented fossil records of biological morphology, robotic systems use reverse engineering to create interactive physical models, allowing scientists to test critical evolutionary hypotheses in controlled environments. Ishida et al.[221] introduced "paleoinspired robotics", an innovative field that builds on existing bioinspired robotics, enabling the study of ancient organisms and their evolutionary trajectories. Crucially, robotic experimental platforms allow for parameter adjustments, such as joint stiffness and motion frequency, enabling researchers to systematically explore how extinct species adapted to environmental changes. Research on the terrestrial adaptation of *Polypterus senegalus* highlights the preadaptive role of phenotypic plasticity in tetrapod land colonization[222]. Their skeletal remodeling and locomotor optimization (reduced motion redundancy, enhanced pectoral fin support) align with early fossil traits, suggesting that developmental plasticity drives morphology-function coevolution through mechanical loading. Inspired by this, bioinspired soft robots can replicate musculoskeletal synergy by using flexible materials that dynamically respond to terrestrial mechanics (ground reaction force-triggered deformation). Soft robotic models with "deformable pectoral fins" emulate aquatic-terrestrial transition strategies, validating evolutionary hypotheses while advancing adaptive soft robotics.

For bioinspired robots, a salamander-inspired robot with central pattern generators (CPGs) replicated the gait of the Devonian tetrapod *Orobates pabsti*. Its movement closely matched fossil footprints, with a 92% congruence, confirming the energy efficiency optimization of early terrestrial locomotion[223]. Fluid dynamics experiments using 3D-printed plesiosaur pectoral fins demonstrated a synergistic vortex effect between the anterior and posterior fins: when the fins were 90° out of phase, the rear fins captured vortex energy from the front, boosting thrust by 40% (**Figure 9A**). This challenges traditional assumptions that long necks create high drag, supporting the hydrodynamic feasibility of the plesiosaur's unique morphology[224]. A robotic tail model of Spinosaurus aegyptiacus, fabricated from a 0.93-mm-thick flexible plate, generated 8.1 times more thrust and 2.6 times greater efficiency than tails of terrestrial theropods, supporting its capacity for aquatic undulatory swimming and a specialized aquatic lifestyle (Figure 9B)[225]. Fossil records of soft tissues are scarce, leading paleontologists to hypothesize about lost tissues based on extant animal structures[226]. For example, a silicone-carbon fiber composite bioinspired prototype of the *Australovenator* foot (reconstructed from scale impressions) revealed that dynamic soft tissue mechanics, rather than skeletal structure alone, governed the diversity of footprints[227].

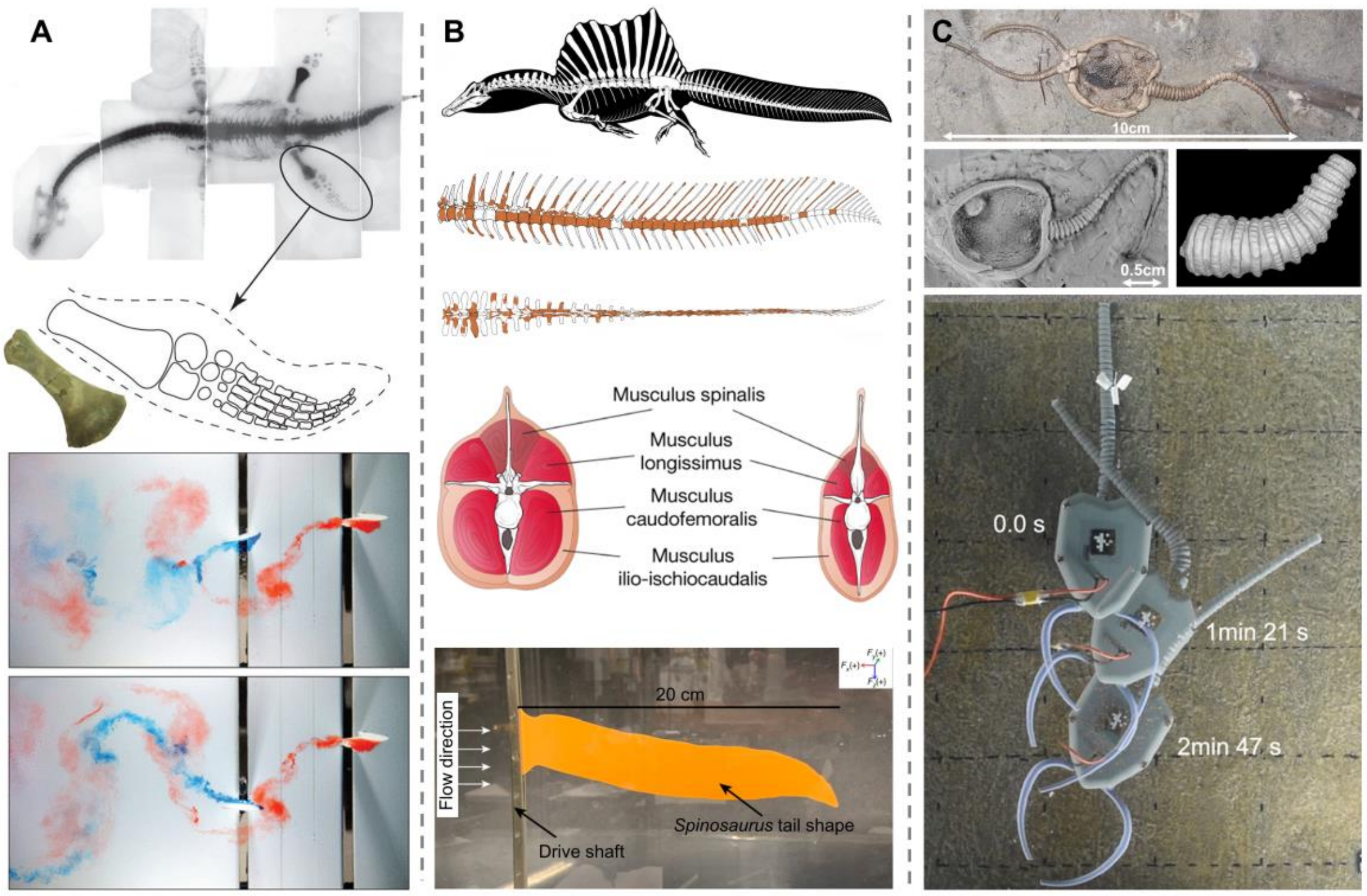

**Figure 9. Representative cases of bioinspired robots for paleontology.** (A) Experimental fluid dynamics of bioinspired plesiosaur flippers indicate that vortex interactions can enhance the efficiency of the hind flippers. Reproduced with permission.[224] Copyright 2017, The Royal Society. (B) A bioinspired robotic tail model of *Spinosaurus aegyptiacus* validated its capability for aquatic propulsion, providing experimental support for functional interpretations of extinct dinosaur locomotion. Reproduced with permission.[225] Copyright 2020, Springer Nature. (C) Bioinspired robots can test locomotion hypotheses of extinct species. By mimicking *Pleurocystitids*, Rhombot demonstrated that wide-amplitude stem oscillations and brachioles-first propulsion significantly improved speed and efficiency, aligning with fossil-based evolutionary trends. Reproduced under the terms of a CC BY-NC-

ND license.[228] Copyright 2023, National Academy of Sciences.

The underwater bioinspired robot Rhombot is an excellent example of using robots to study extinct marine organisms. By mimicking the extinct echinoderm *pleurocystitids*, Rhombot demonstrated that wide-amplitude stem oscillations, combined with brachioles-first propulsion, allowed it to move nearly a hundred times faster than traditional stem-first motion, greatly improving energy efficiency (Figure 9C). The optimal stem-to-body length ratio of 3 for maximum speed matches evolutionary trends seen in fossils, suggesting that natural selection favors morphological changes that improve movement efficiency[228]. By bridging paleontology and engineering, bioinspired robots provide an empirical platform for testing evolutionary drivers, converting speculative theories about morphology into experimentally validated biomechanical principles.

## 5. Applications of bioinspired underwater soft robots

Bioinspired underwater soft robots provide significant advantages in flexibility, adaptability, and safe interaction with delicate environments. These robots, designed to mimic the efficiency and agility of marine animals, can navigate through complex underwater terrains and execute tasks that traditional rigid robots struggle with. Their applications include underwater manipulation, exploration, and medical fields.

### 5.1 Underwater manipulation

Soft robots excel in handling fragile objects due to their inherent compliance, eliminating the need for extensive sensorization. Underwater soft robotic manipulation is primarily achieved through grasping or adhesion. Grasping involves force-based holding, exemplified by finger-inspired soft grippers capable of handling coral (**Figure 10A**)[229], jellyfish (Figure 10B)[230], sea cucumbers[231], and ancient porcelain[232]. Suction adhesion, driven by pressure differences, enables attachment to surfaces. Inspired by the remora's sucker disc, Li et al.[233] designed a biomimetic adhesion device that can achieve redundant and adaptive adhesion in water and can also be used for applications such as hitchhiking (Figure 10C). However, the application of suction-based grippers is often limited to objects with flat or nearly flat surfaces. Addressing this, Xie et al.[77] and Wu et al.[234] developed bioinspired octopus arm-like grippers combining suction and grasping, allowing efficient manipulation of objects of varying shapes and sizes (Figure 10D and 10E).

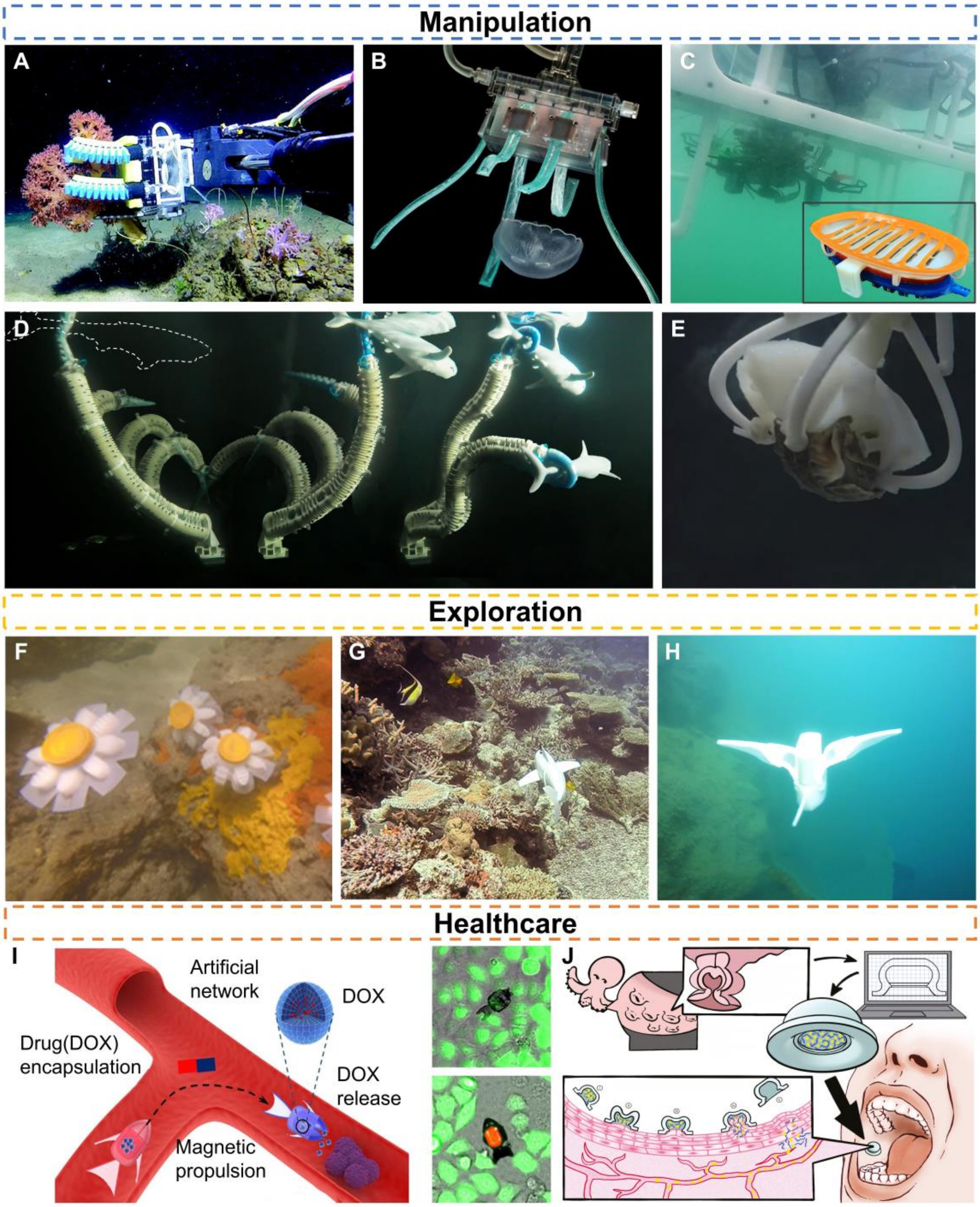

**Figure 10. Underwater applications of bioinspired soft robots.** Manipulation: (A) Deep-sea soft gripper. Reproduced under the terms of a CC-BY license[229] Copyright 2016, Mary Ann Liebert, Inc. (B) Ultra-soft gripper. Reproduced with permission.[230] Copyright 2019, AAAS. (C) Remora-inspired suction device. Reproduced with permission.[233] Copyright 2022, AAAS. (D) Bioinspired octopus-arm soft gripper combining suction and grasping. Reproduced with permission.[77] Copyright 2023, AAAS. (E) Self-Adaptive grasping. Reproduced under the terms of a CC-BY license.[234] Copyright 2022, Wiley-VCH. Exploration: (F) Soft robotic jellyfish.

Reproduced with permission.[235] Copyright 2018, IOP Publishing. (G) SoFi robotic fish. Reproduced with permission.[236] Copyright 2018, AAAS. (H) Deep-sea snailfish-inspired robot. Reproduced with permission.[237] Copyright 2021, Springer Nature. Healthcare: (I) Shape-morphing microfish. Reproduced with permission.[132] Copyright 2021, American Chemical Society. (J) Octopus-inspired suckers. Reproduced with permission.[238] Copyright 2023, AAAS.

### 5.2 Underwater exploration

Soft robots offer unique locomotion capabilities, surpassing rigid robots in navigating complex underwater environments. Inspired by marine life, researchers have developed soft robots for robust and controlled underwater movement. Frame et al.[235] created a bioinspired soft robotic jellyfish that swims untethered off Florida's Atlantic coast, showcasing its potential for low-impact exploration of coral reefs and complex underwater structures like shipwrecks and pipelines (Figure 10F). Robert et al.[236] introduced SoFi, a soft robotic fish with acoustic communication for close-range marine life observation and disturbance-free underwater filming (Figure 10G). Li et al.[237], inspired by the deep-sea snailfish, engineered an untethered soft robotic fish capable of deep-sea exploration (Figure 10H). This robot performed 45-minute flapping actuation tests at 10,900 meters in the Mariana Trench and swam at 5.19 cm/s at 3,224 meters in the South China Sea, demonstrating the potential for deep-sea biology, geological studies, and environmental monitoring.

### 5.3 Medical applications

Bioinspired soft robots have shown promise in biomedical applications, particularly in targeted drug delivery and non-invasive therapeutic interventions. These robots can navigate through the vasculature and release drugs in a controlled manner using external stimuli such as magnetic fields, ultrasound, or temperature. Shape-morphing microfish (SMMF), fabricated via 4D laser printing with pH-responsive hydrogels and magnetic nanoparticles, combine magnetic actuation with adaptive shape transformation (Figure 10I)[132]. The SMMF encapsulates anticancer drugs (doxorubicin) by closing its mouth at physiological pH (7.4) and releases the payload in tumor microenvironments (pH <7), achieving precise HeLa cell destruction, with experiments showing an 80% HeLa cell mortality rate within 6 hours. Luo et al.[238] developed an octopus sucker-inspired buccal patch that enhances systemic absorption of peptide drugs through a synergistic mechanism of mechanical stretching and chemical permeation enhancers (Figure 10J). Mimicking the bilayered structure of octopus suckers, the patch applies negative pressure to expand the buccal mucosa while using permeation enhancers such as sodium taurocholate to disrupt epithelial barriers. This strategy significantly improved the bioavailability of desmopressin and demonstrated safety and tolerability in human clinical trials.

Bioinspired underwater soft robots have demonstrated remarkable versatility across a wide range of applications, from underwater manipulation, exploration to specific medical applications. Their compliance and adaptability allow for safe interaction with delicate marine ecosystems, while their ability to navigate complex aquatic environments surpasses that of traditional rigid robots.

## 6. Future research directions toward biouniversal-inspired robotics

For the future of biomimetics, we propose the nascent field of biouniversal-inspired robotics, which shifts beyond single-organism studies to uncover broader biological patterns and the universal principles governing them. These principles are often rooted in convergent evolution, where organisms from distinct evolutionary clades—spanning broad taxonomic ranks and lacking the most recent common ancestor (MRCA)—develop similar morphologies or structural features in response to shared environmental stress and ecological demands[239].

This phenomenon can be understood as the result of ecological niche shaping, where a selective pressure matrix, comprising environmental factors such as light, temperature, salinity, and water flow rate, acts like a sieve or mold, driving different species toward similar functional phenotypes. Simultaneously, genetic constraints limit evolutionary possibilities, further shaping adaptive trajectories[240]. Convergent evolution is widespread in nature and can explain similar adaptations across diverse biological groups.

Convergent evolution has occurred repeatedly across diverse lineages throughout the tree of life. In plants, the need for defense and water conservation has driven different lineages to independently evolve similar thorn-like structures[241]. In animals, convergent evolution is even more diverse. Flying fish (Actinopterygii), pterosaurs (Reptilia), birds (Aves), and bats (Mammalia) have all independently developed aerodynamically similar wing structures from distinct evolutionary origins to expand aerial ecological niches or evade predators (**Figure**

**11A**)[242–244]. Likewise, in aquatic environments, diverse lineages have independently evolved fin-like structures for efficient propulsion. Highly specialized marine birds such as penguins, marine mammals represented by whales and dolphins, and large extinct marine reptiles such as ichthyosaurs have all convergently evolved similar flipper-shaped morphologies adapted for swimming—despite profoundly divergent evolutionary origins in their appendages (a distinction profoundly reflected in the internal anatomical structures of their flippers, see Figure 11B)[245,246]. These convergent fin morphologies vividly demonstrate how biomechanical demands of aquatic locomotion sculpt similar adaptations across distantly related vertebrate lineages.

In parallel, many aquatic organisms have independently evolved suction-based attachment mechanisms to cope with strong water currents and complex substrates (Figure 11C)[247–249]. Sea lampreys use circular oral disks with keratinized tooth-like structures; blephariceridlarvae employ abdominal suction pads; clingfish form ventral suction cups from modified pelvic fins; and *Spirula spirula* utilizes microstructured arm suckers. Though these systems are anatomically diverse, they share friction-enhancing features and high surface compliance, reflecting convergent functional solutions for temporary yet robust adhesion in hydrodynamically challenging environments.

One of the most striking examples of convergent evolution is the hydrodynamic adaptation of aquatic mammals and fish. Despite their entirely different evolutionary origins—whales, for example, evolved from terrestrial artiodactyls that returned to the ocean around 50 million years ago, whereas fish have occupied aquatic ecosystems for over 500 million years—both groups have independently evolved highly streamlined body shapes to minimize drag and optimize swimming efficiency[240,250]. These cases exemplify how species subjected to similar environmental pressures or functional demands develop analogous morphologies and functionally convergent adaptations as survival strategies. This convergence pattern highlights how ecological niches shape species by acting as an evolutionary mold. It also suggests that repeatedly evolved morphological and functional traits, such as spines, fin/wing structures, adhesion organs, and streamlined bodies, transcend the genetic constraints of individual species and exhibit universal adaptability to environmental pressures. These convergent phenotypes, driven by physical constraints toward optimal engineering solutions, provide inspiration for biouniversal design across scales and mediums.

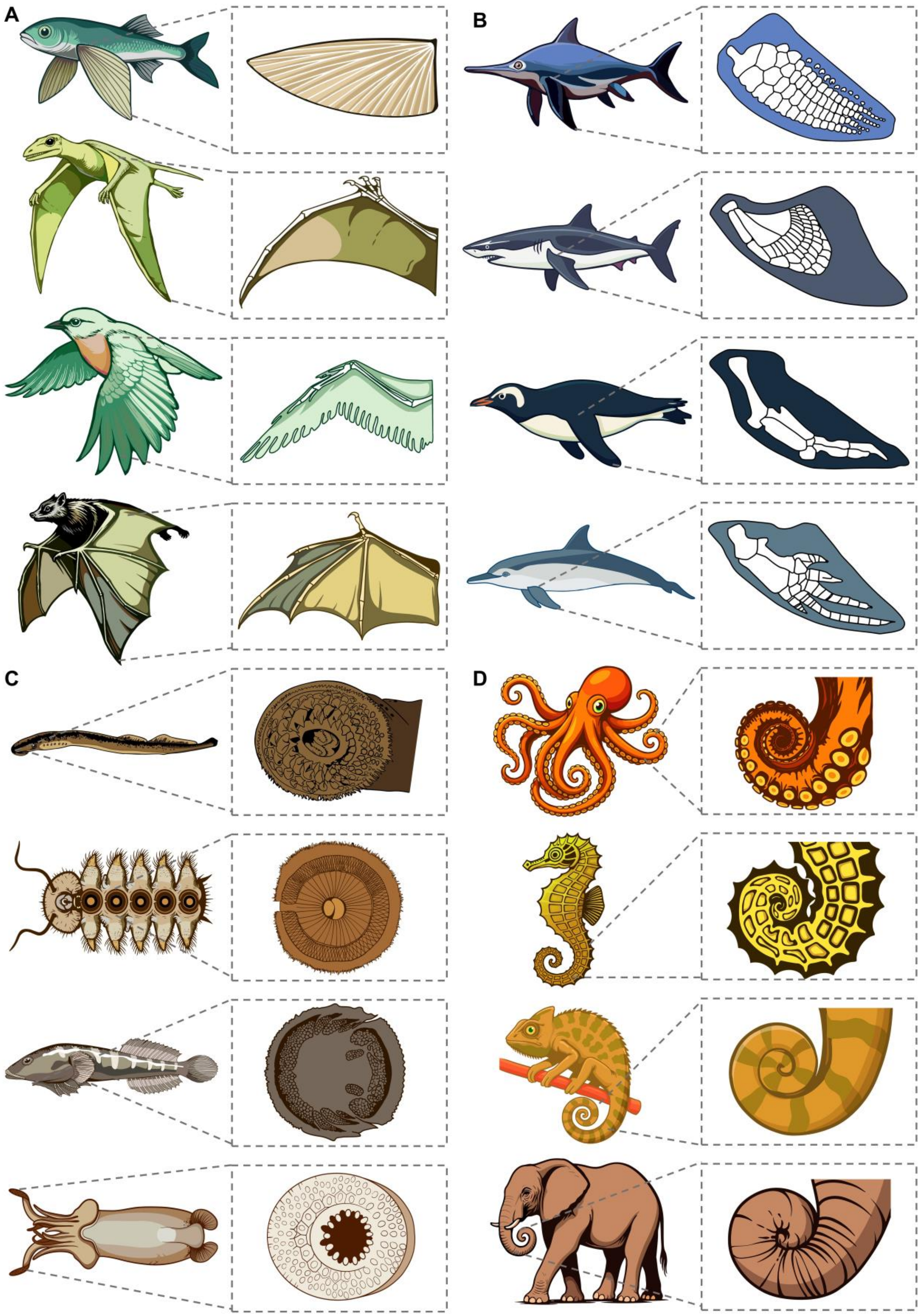

**Figure 11. Convergent evolution across diverse species driven by similar functional demands.** (A) Powered flight: Despite vast evolutionary distances, flying fish, pterosaurs, birds, and bats have converged on wing-like structures that enable aerial locomotion, illustrating how similar aerodynamic constraints shape structural evolution. (B) Aquatic locomotion: Distant vertebrate lineages—including extinct reptiles (ichthyosaur), mammals (whale, dolphin), and birds (penguin)—have independently evolved streamlined body shapes and fin-like appendages to enhance swimming efficiency in marine environments. (C) Aquatic attachment organs: Suction-based adhesion structures have independently evolved in organisms such as sea lampreys, blephariceríd larvae, clingfish, and the deep-sea squid *Spirula spirula*. These organs often exhibit micro- and nano-scale surface textures that enhance friction and interlocking with substrate asperities, providing strong attachment in high-flow environments. (D) Logarithmic spirals in body structures: Animals as diverse as octopuses, seahorses, chameleons, and elephants have evolved spiral-shaped appendages (arms, tails, trunks) that follow logarithmic geometry, suggesting functional convergence in grasping, coiling, and compact storage. Modified from Wang et al., 2024[251].

Research on biouniversal-inspired robotics remains in its early stages. Wang et al.[251] compared representative species with grasping appendages from both vertebrates (such as seahorses, chameleons, and elephants) and invertebrates (such as octopuses). Despite vast differences in anatomical structures (e.g., skeletal presence, muscle architecture) and environmental adaptations (aquatic vs. terrestrial), these organisms all exhibit a common logarithmic spiral pattern in their grasping structures (Figure 11D). This cross-species universal grasping mechanism was mathematically modeled and translated engineering design principles for robotics. By discretizing the spiral structure and employing 3D printing with thermoplastic polyurethane (TPU) materials, along with dual- or triple-cable antagonistic actuation, they developed bioinspired soft robots capable of adaptive grasping with rapid, scalable fabrication across size ranges from centimeters to meters. While this work successfully demonstrated a functional prototype based on the logarithmic spiral, it raises a deeper biological question: why has convergent evolution across such taxonomically distant species favored this specific geometric pattern? Despite decades of research, the biological rationale behind the prevalence of the logarithmic spiral in grasping structures remains unresolved. Mechanical modeling and experimental robotic studies may offer new insights into the functional advantages and evolutionary origins of this form. In this regard, biouniversal-inspired robots not only serve engineering purposes, but also provide a powerful, controllable platform for probing universal principles in biology through physical embodiment and hypothesis-driven experimentation.

**7. Conclusion**

This review establishes a comprehensive framework for bioinspired underwater soft robotics, integrating biological principles from materials, morphology, kinematics, sensing, and swarm strategies into robotic design and implementation. It highlights the bidirectional flow of knowledge, where biological adaptations drive robotic advancements, and soft robotic platforms, in turn, serve as experimental tools to investigate biomechanical and evolutionary principles. Notably, soft robots offer unique advantages in reconstructing and validating paleobiological locomotion, providing unprecedented insights into extinct species' functional morphology and movement dynamics. Soft robots have demonstrated superior adaptability in underwater manipulation, exploration, and medical applications, outperforming traditional rigid systems in complex and dynamic environments. Moreover, we outline future research directions toward biouniversal-inspired robotics, an emerging paradigm that transcends single-species studies to uncover the universal principles shaped by convergent evolution across diverse taxa.

However, significant challenges remain on both biological and robotic fronts. From a biological perspective, biomimetic studies often concentrate on individual species, limiting the discovery of universal principles. Many complex biological features, such as multi-scale adaptive responses and nonlinear control strategies, remain difficult to decipher and replicate in robotic systems. From a robotics perspective, soft robots face persistent challenges in material durability under extreme underwater conditions, limiting their long-term deployment. The fabrication of complex multi-material structures remains technically demanding, hindering scalability. Limited energy constrains operational duration, while real-time sensing and adaptive control capabilities lag behind biological systems, reducing their responsiveness in dynamic environments. Moreover, the coordination and environmental adaptability of swarm intelligence algorithms remain unresolved. Addressing these challenges requires a more integrative, cross-species approach, along with advances in bioinspired materials, energy-efficient actuation, multi-modal sensing, and AI-driven control strategies to improve both biological realism and the functional performance of underwater soft robots.

Bioinspired underwater soft robotics holds immense potential to transform marine science and industry. These robots could enable unprecedented exploration of uncharted deep-sea ecosystems, monitor climate-sensitive coral reefs with minimal disturbance, and deliver targeted therapies in minimally invasive medical procedures. Beyond practical applications, they deepen our understanding of evolutionary biology and inspire public engagement with marine conservation. However, ethical considerations must be carefully addressed as the field advances. These include potential ecological disruption from the prolonged deployment of robotic systems in sensitive marine habitats, ethical concerns surrounding the use of living tissues in biohybrid constructs, and unintended behavioral disturbances to wildlife caused by artificial presence or acoustic signals. A commitment to responsible innovation—grounded in ecological awareness and ethical stewardship—will be critical for developing sustainable technologies that coexist harmoniously with fragile marine ecosystems.

The journey from biology to robotics and back is not merely a technical endeavor but a shift in our scientific perspective on learning from and contributing back to nature. By embracing biomimicry as both a tool and a teacher, we not only advance science and technology, but also honor the deep wisdom embedded in nature. We carry the responsibility to let our creations coexist with the ecosystems that inspired them—guiding robotics not only toward innovation, but toward harmony, humility, and stewardship.